\documentclass{article}

\usepackage{PRIMEarxiv}
\usepackage[symbol]{footmisc}

\usepackage[utf8]{inputenc} 
\usepackage[T1]{fontenc}    
\usepackage{hyperref}       
\usepackage{url}            
\usepackage{booktabs}       
\usepackage{amsfonts}       
\usepackage{nicefrac}       
\usepackage{microtype}      
\usepackage{lipsum}
\usepackage{fancyhdr}       
\usepackage{graphicx}       
\graphicspath{{media/}}     
\usepackage{tikz}
\usepackage{amsmath}
\usepackage[toc]{appendix}
\usepackage{amssymb}
\newcommand{\A}{\mathcal{A}}

\newcommand{\State}{\mathcal{S}}

\newcommand{\Hist}{\mathcal{H}}
\newcommand{\ind}{1}
\newcommand{\Ell}{\mathcal{L}}

\newcommand{\Nz}{\mathbb{N}_0}

\usepackage[belowskip=-15pt,aboveskip=0pt]{caption}
\usepackage{enumitem}
\newlist{enumsteps}{enumerate}{1}
\setlist[enumsteps,1]{leftmargin=20mm, wide=0pt, label=Assumption \arabic* }

\author{
  Conor M. Artman \thanks{Denotes first authorship.} \\
  Machine Learning Group \\
  Lawrence Livermore National Laboratory \\
  Livermore, CA\\
  \texttt{artman1@llnl.gov} \\
   \And
    Aditya Mate \thanks{All contributions were made while interning at IBM Research} \hspace{.25mm} \thanks{Denotes equal contribution.} \\ 
    Microsoft New England \\
    Cambridge, MA
    \And
    Ezinne Nwankwo \footnotemark[3] \\
    University of California, Berkeley \\
    Berkeley, CA
    \And
    Aliza Heching \\
    \textbf{Tsuyoshi Id\'e} \\
    \textbf{Ji\v{r}\'\i\, Navr\'atil} \\
    \textbf{Karthikeyan Shanmugam}
    \thanks{All work was completed while at IBM Research.} \\
    \textbf{Wei Sun} \\
    \textbf{Kush R. Varshney} \\
    IBM Research -- T.J. Watson  \\
    Yorktown Heights, NY
    \And
    Lauri Goldkind \\
    Fordham University \\
    Bronx, NY
    \And 
    Gidi Kroch \\
    Leket Israel \\
    Ra'anana, Israel
    \And 
    Jaclyn Sawyer \\
    Breaking Ground \\
    New York, NY
    \And
    Ian Watson \\
    Change Machine \\
    Brooklyn, NY
}

\title{A Resource-Constrained Stochastic Scheduling Algorithm for Homeless Street Outreach and Gleaning Edible Food}

\begin{document}
\maketitle

\begin{abstract}

We developed a common algorithmic solution addressing the problem of resource-constrained outreach encountered by social change organizations with different missions and operations: Breaking Ground---an organization that helps individuals experiencing homelessness in New York transition to permanent housing and Leket---the national food bank of Israel that rescues food from farms and elsewhere to feed the hungry. Specifically, we developed an estimation and optimization approach for partially-observed episodic restless bandits under $k$-step transitions. The results show that our Thompson sampling with Markov chain recovery (via Stein variational gradient descent) algorithm significantly outperforms baselines for the problems of both organizations. We carried out this work in a prospective manner with the express goal of devising a flexible-enough but also useful-enough solution that can help overcome a lack of sustainable impact in data science for social good.

\end{abstract}


\maketitle
\newpage
\section{Introduction}

\renewcommand*{\thefootnote}{\arabic{footnote}}

Stochastic scheduling methods have been applied to problems arising in several application domains that serve the social good, such as medical treatment adherence (tuberculosis and sleep apnea), maternal and newborn health information dissemination, hepatitis C treatment prioritization, disease screening (carcinoma and COVID-19), and anti-poaching, with solutions using multi-armed bandit algorithms \citep{ayer_prioritizing_2019,herlihy_planning_2022, lee_optimal_2019, mate_collapsing_nodate,marecek_screening_2020,mate_field_2021,shi2022bandit}. In this paper, we report on our work addressing similar resource-constrained scheduling problems faced by two social change organizations: Leket, the national food bank of Israel, and Breaking Ground, a provider of homeless outreach in New York City. 

Breaking Ground (BG) is a social services organization in New York City that provides high-quality permanent and transitional housing for people experiencing homelessness. Its philosophy holds that supportive housing is of substantially lower cost than homeless shelters---and many times lower cost than jail cells or hospital rooms, and that people with psychiatric and other problems can better manage those problems once they are permanently housed and provided with services. Since its founding in 1990, BG has provided 5,000 housing units. Since 2000, BG has expanded to develop 15 more residences with 1,000 units in various stages of development.

With many individuals experiencing homelessness in New York City in need  of social services, Breaking Ground has a dedicated team of outreach workers who traverse the streets daily and nightly to meet these people and offer help. The organization's primary goal is to transition their clients into permanent housing. Clients may be in several progressively improving states: on the street, in stabilization housing, in transitional housing, or in permanent housing. Building trust is an important part of moving individuals experiencing homelessness along toward more permanent housing solutions. BG is constrained by a limited number of outreach workers. The main action is engaging with a given client on a given day (or not). Even if an outreach worker wants to contact a client, there is uncertainty because the client may not be found where they were last contacted. Thus, a critical problem for BG is planning when and where to send outreach workers with the goal to keep individuals experiencing homelessness progressing among states of housing, which can be modeled as a resource-constrained stochastic scheduling problem and solved with a multi-armed bandit algorithm. Homeless clients can move among states of housing without having had an explicit outreach interaction, which implies the use of \emph{restless} multi-armed bandit (RMAB) algorithms.

\setcounter{footnote}{0} 

Leket is the largest food rescue organization in Israel.\footnote{Leket's name derives from the Torah's description of ``gleaning''. Gleaning in this sense refers to donating ears of grain to impoverished people if they fall from a reaper's hand while harvesting (among other agricultural donations).} For some farmers, harvesting all their produce may be unprofitable, while others lack the resources to harvest their crop before putrefaction. In either case, tens of millions of pounds of fresh fruits and vegetables may be wasted each year, so Israeli farmers invite Leket to harvest and redistribute this produce to people in need around the country. Leket also proactively scouts farmers with unharvested produce year-round. Leket coordinates with farmers to pick up, store, and distribute excess food to 175,000 people weekly. Leket also prepares and delivers 4,500 breakfasts to students across $70$ schools in Israel.  

The basic problem for Leket is also one of resource-constrained stochastic scheduling. Time-constrained Leket staff members call farmers to scout if there is edible food to be gleaned. There is uncertainty in this outreach because farmers may be unreachable in their fields or too busy to respond. The state of the farm, whether there is or is not produce available, is only observed if contact is made with the farmer, which is equivalent to activating the arm corresponding to that farm in an RMAB. 

We reached the problem specification for the two organizations---including states, actions, constraints, and what is observed---after significant dialog between the subject-matter experts and the technical experts among the authors, which we feel is itself a key contribution.  As we discuss next, a common problem specification permits us a reusable algorithm that overcomes the lack of sustainable impact seen in the area of data science for social good.

While the data science for social good movement has generated significant progress toward harnessing AI technologies for social good organizations working in food security, poverty, and other socioeconomic inequities, nearly all of these projects have failed to move beyond the initial proof-of-concept \citep{shi_artificial_2020, varshney_open_2019,  chui_michael}. Consequently, few have managed to have sustained impact. This is caused, in part, by insufficient technical resources for integrating and maintaining bespoke solutions across social change organizations. For example, many solutions are often started as pro bono work by data scientists who then pass off a solution with insufficient infrastructure to sustain it, which further exacerbates this issue \citep{chui_michael,varshney_open_2019,shi_artificial_2020}. We believe that this has emerged as a trend because committing to the \underline{pro}spective development of reusable assets at the outset may appear risky: before the collaboration between experiential/subject-matter and technical experts is under way, the technical experts must \textit{assume} there is nontrivial crossover across disparately related social change organizations.\footnote{The IBM Research authors sought out Leket and Breaking Ground as collaborators precisely because we anticipated the organizations would require solutions involving resource-constrained stochastic scheduling, which can be addressed using RMAB algorithms. A third organization involved in this prospective work, Change Machine, has a somewhat similar problem, but more complicated states and actions for which we were unable to develop a common formulation.}

One way to address the lack of sustainable impact is to develop generic and reusable technical assets that can be maintained and delivered as a service \citep{varshney_open_2019}. Possibilities include (1) a widely-applicable algorithm that meets the needs of many social change organizations, (2) a comprehensive toolkit of methods that applies to numerous social change problems (e.g., AI Fairness 360 \citep{bellamy_ai_2018, richardson_framework_2021}), or (3) a pre-trained foundation model. In this work, we choose the first possibility. We note that while \citet{shi2022bandit} pose a single algorithm for food rescue and anti-poaching, it is only supported by retrospective simulation studies. What we attempt to do prospectively with a single algorithm is unique in data science for social good.

\section{Problem Formulation}\label{sec:method}

We now set forth the problem setup and solution requirements for the scheduling goals faced by Leket and BG. In the real-world settings, states are partially observed (which is equivalently described as partially monitored arms of the RMAB) and state transitions may have gaps (which is called $k$-step transitions) because we only get to know the housing status or availability of food to glean when outreach is successful. There are four states in the BG case (street, stabilization, transitional, permanent) and more than two states in the general case. Also for technical reasons discussed in Appendix \ref{sec:rel_work}, the RMAB policy must be interchangeable with any deterministic policy, rather than focusing only on the commonly-used \textit{Whittle index policy} (see Appendix \ref{wrelax} for the Whittle index policy's technical definition and derivation). Given these considerations, we require:
\begin{enumerate}
    \item Estimation of transition dynamics from partially observed RMAB arms \textit{and} $k$-step transitions.
    \item Flexibility to accommodate policies outside of Whittle index policy approaches (see Appendix \ref{sec:rel_work} for further discussion).
    \item Computationally tractable and valid estimation for per-arm state spaces larger than $2$.
\end{enumerate}
To the best of our knowledge, there is no RMAB algorithm in the online learning setting that can practically accommodate all of these considerations simultaneously, with or without approximate Bayesian inference. In this section, we step through each of these requirements and motivate our choices for our final algorithm. From there, we state assumptions on our model before proceeding to formal algorithmic descriptions and experiments.

\subsection{Learning Problem: Partial-Observability and $k$-Step Trajectories in RMABs}\label{subsec:learning}

\begin{figure}
    \centering
    \includegraphics[width=.7\linewidth, height =.4\linewidth]{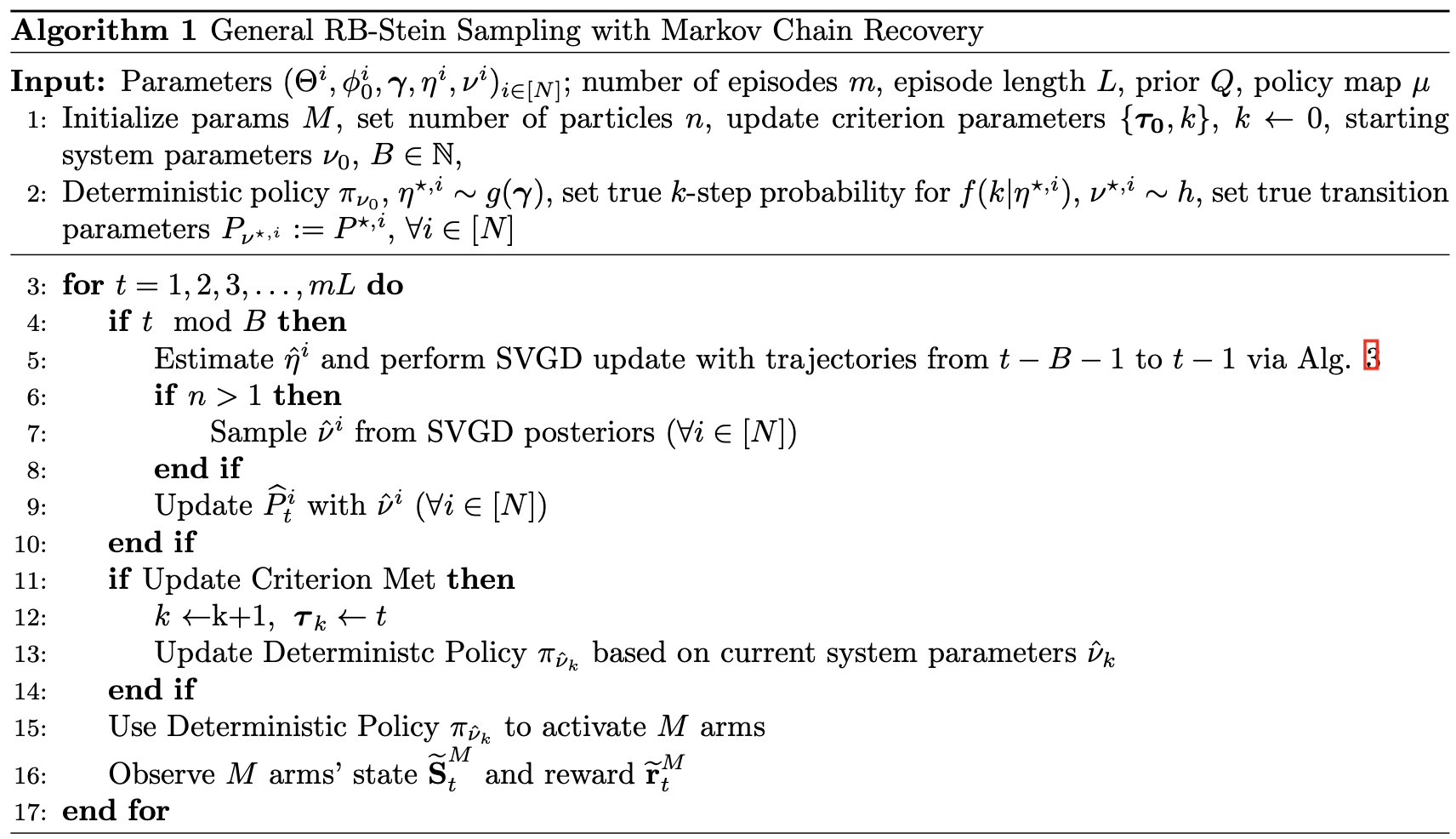}
    \caption{Restless Bandit Stein Sampling (RB-Stein)}
    \label{alg:gen_alg}
\end{figure}

Consider states $x,y,z \in \State$ and let state transitions from, say, $x$ to $y$ to $z$ be denoted $x \xrightarrow[]{} y \xrightarrow[]{} z$. For $N$ arms in the RMAB, we wish to estimate transition matrices $P^{a}_{i}$, $a \in \A$, where $(P^{a}_{i})_{xy} = p^{i}(y|x,a)$, $(\forall i \in [N])$. Our RMAB must handle two issues as it chooses actions:

\begin{enumerate}
    \item \textbf{$k$-Step Unlabeled Transitions}. An unlabeled transition refers to the scenario where we observe a state at time $t$ and then the \textit{number} of transitions between some time $t$ and $t+(k+1)$ in the data, but we do not observe those particular realizations of the state. This scenario has been referred to as the case of ``unlabeled'' state transitions \citep{kohjima_learning_2020}. For example, if the true underlying state transition were $x \xrightarrow[]{} y \xrightarrow[]{} z$, a 1-step unlabeled transition could appear in our data as $x \xrightarrow[]{} ? \xrightarrow[]{} z$, $? \xrightarrow[]{} y \xrightarrow[]{} z$, or $x \xrightarrow[]{} y \xrightarrow[]{} ?$, depending on when we started or stopped observing data. For a  $2$-step unlabeled transition that starts at some state $w$, one realization could be $w \xrightarrow[]{} ? \xrightarrow[]{} ? \xrightarrow[]{} z$, $? \xrightarrow[]{} x \xrightarrow[]{} ? \xrightarrow[]{} z$, and so on. We refer to a transition with a finite number of unlabeled transitions $k$ as a $k$-step unlabeled transition, but in general we will refer to these as $k$-step transitions. 
    \item \textbf{Partial Observability}. When we only observe some subset $M < N$ of the arms each time-step, where $M$ is the RMAB's budget constraint, the RMAB is called \textit{partially observed}. The states and rewards of the other arms are not observed.\footnote{Classic RMABs assume \textit{fully observed} states for every arm. This means that for every time-step, every arm's state and corresponding reward is observed.} This is also sometimes called, \textit{partial monitoring}.

\end{enumerate}

\noindent
Taking $(1)$ and $(2)$ together, in our problem setting we only observe $M$ out of $N$ arms at any time $t$ and our data takes the form of $k$-step trajectories.

\emph{Extension from MCs to MDPs.} We start our development from \citet{kohjima_learning_2020}. Let $K_s$ be the random variable determining the realization of $k$ at data-generating time $s$ for a $k$-step transition. $K_s$ is modeled as an exponential family model $f$ parameterized by $\eta \in \mathbb{R}^{|\State|}$. 
\[
P(K_s = k | X_s = i) = f(k|\eta_i)
\]
\noindent
Given a transition matrix $P$, a $k$-step transition probability is given by $P^k$, e.g., $P^2$ yields $2$-step probabilities and $P^{15}$ yields $15$-step probabilities \citep{durrett_probability_2019}. Therefore, the probability of moving from state $i \xrightarrow[]{} j$ after evolving a MC $k$ steps is $(P^k)_{ij} = P(X_{s+1} = j | X_s = i, K_s = k )$. Together, this yields
\[
P(X_{s+1} = j, K_s = k | X_s = i ) = f(k | \eta) P^{k}_{ij}.
\]
\noindent

Set $\theta := \{\eta, \nu\}$ and let $M_{ijk}$ be the number of transitions made from $i \xrightarrow[]{} j$ in $k$ steps. By \citet{kohjima_learning_2020}, the likelihood function for unlabeled transition data is

\begin{equation}\label{eqn:like}
    \ell(\theta) = \sum_{i,j,k} M_{ijk} \log [f(k|\eta_{i}) (P_{\nu})^{k}_{ij}]
\end{equation}

With this likelihood, the next step is to develop a posterior sampling approach. While \citet{kohjima_learning_2020} offer an algorithm for handling unlabeled transition data in Markov chains (MCs), our setting requires an extension to MDPs in the RMAB framework. We handle this in two steps. 

First, we integrate the learning problem with the MDP framework by considering augmented state-action pairs as our states. We collect data under a fixed policy and work with augmented states to preserve the Markov property, denoted $Z_t:=(X_t, A_t) \in \mathcal{Z}$, where $\mathcal{Z}:=\State \times \A$. We use these states in our likelihood in Equation \ref{eqn:like}, so that our transition parameters are stored in a $|\mathcal{Z}| \times |\mathcal{Z}|$ transition matrix now rather than $|\State| \times |\State|$. In contrast to \citet{kohjima_learning_2020}, who estimate their parameters via a Maximization-Minimization algorithm, we sample parameters from a posterior distribution approximated by SVGD. Finally, we also must ensure any fixed policy sufficiently samples $(X_t, A_t)$ pairs, so we incorporate an episodic update criterion into our algorithm. Algorithm $\ref{alg:gen_alg}$ steps through our estimation procedure. 

\emph{Planning Problem: Deterministic and Index Policies.}
While our algorithm accommodates any deterministic policy, we briefly review the special case of index policies, as they are by far the most common in RMABs and can be optimal under special circumstances \citep{nino-mora_markovian_2023}. 

In general, using an index policy entails maintaining an index value for each arm at each time step. The idea is to exploit the arm that appears most promising based on its current index. The primary differences among index policies are what data they use to estimate indices and to what degree they attempt to measure the long-term changes in each arm's reward distribution. In the simplest case of the \textit{myopic} policy, also known as the ``myopic index policy'' or the ``index heuristic policy'', each arm's index measures the arm's potential for high rewards in the next immediate time step without conditioning on the arm's future potential for rewards. So, at each time step, the myopic policy selects the arm with the highest index value. Examples of other commonly used index rules include the Gittins index policy, Whittle index policy, and knowledge gradient index policy \citep{cochran_knowledge_2011, gittins_bandit_1979, whittle_restless_1988}. 

When the RMAB is indexable, index policies other than the WIP are usually suboptimal but the non-trivial problem of establishing the indexability condition can be a major impediment for real-world applications \citep{nino-mora_markovian_2023}. One line of work explores simpler alternatives, while another explores another index policy with fewer technical requirements \citep{wang_optimality_2012, sombabu_whittle_2020, xiong_index-aware_2022, xiong_model-free_2022}.

In the former case, the primary candidate is the myopic policy. The myopic policy is usually used as a baseline due to its simplicity and computational efficiency, making it useful for real-time decision-making in dynamic environments. However, it may not always achieve the best possible long-term performance since it disregards the future consequences of actions. More sophisticated policies, such as look-ahead or learning-based strategies, e.g., Thompson sampling, can be employed to improve upon the myopic approach in restless bandit problems. 

In review, there are nuanced and important context-specific considerations to balance when choosing among index policies. Nonetheless, all of the approaches above are special cases of deterministic policies, so to accommodate potentially \textit{any} of these policies, we do \textit{not} impose any specific deterministic policy in our algorithm development or in its performance analysis. However, as the WIP is a scalable solution and the predominant approach to the RMAB problem at the time of writing, we consider it alongside other common deterministic policies. 

In Section~\ref{data:design}, we first work with synthetic data with fully observed but possibly unlabeled states to study if our integration works well. Second, we extend our RMAB to handle partially observed states and $k$-step transitions. In Section \ref{single_particle:msvgd}, we study performance of our algorithm under a single particle per arm and observe strong performance.

\subsection{Assumptions} We state key assumptions in the RMAB model and then comment on their feasibility. 
\begin{enumsteps}
    \item\label{binassn} (\textit{Binary Action Spaces}) $\A = \{0,1\}$. \\
    
    \item\label{indassn}  (\textit{Indexability}) Denote the set of passive states by $\State_0$ and the set of active ones by $\State_1$. We say an arm is \textit{indexable} when $\State_0$  increases monotonically from $\emptyset$ to the $\State$ as the cost of activating it monotonically increases from $-\infty$ to $\infty$. If every arm $i \in [N]$ is indexable, the RMAB is said to be indexable. See Section $\ref{wrelax}$ for more details on indexability. \\

    \item\label{monassn}  (\textit{Monotone Treatment Effects})
    We assume that the active action, $a = 1$, \textit{always} provides a positive treatment effect relative to the passive action, $a=0$. Concretely, this means that choosing $a=1$ yields an MDP with higher expected rewards than under $a=0$ for all $t$. \footnote{Put another way, applying $a=1$ to every arm would produce the highest possible rewards for all arms in the absence of a budget constraint.} \\

    \item\label{indptassn}  (\textit{Independent Arms})
    Actions on one arm do not affect another arm and rewards of one arm are independent of another arm's rewards. \\

    \item\label{bdkstep} (\textit{Bounded $k$-Step Transitions}) The maximum number of consecutive $k$-step transitions is bounded. Precisely, $\exists k_{\max} \in \mathbb{N}$ such that for any trajectory $\tau$, 
    \begin{displaymath}
        \sum_{s \in \tau} 1 (\{s \hspace{1mm}\text{unlabeled}\}) \leq k_{\max}.
    \end{displaymath} \\
\end{enumsteps}
\vspace{-1.5pc}

While \ref{binassn} is restrictive, in practice it is common to work with a binary action space \citep{avrachenkov_learning_2019}. \ref{indassn} is nontrivial and difficult to justify in practice. Additionally, even if the RMAB is indexable, it is usually computationally expensive to verify. \ref{monassn} is reasonable for applications where there is only one obvious intervention to give, such as gleaning food or interacting with a homeless client. Otherwise, one may consider other methods such as multi-action RMABs, e.g., \citep{killian_robust_2021}. \ref{indptassn} is also mild for the applications where RMABs are natural: farms and homeless clients do not influence each other. Note that if it is not feasible to model each arm as an independent MDP, RMABs may fail to perform well, as most optimization methods for RMABs heavily rely on \ref{indptassn} for convergence guarantees and computational tractability \citep{avrachenkov_whittle_2021, killian_robust_2021, whittle_restless_1988}.  \ref{bdkstep} is necessary to ensure that the worst-case trajectory may have only sparse state information within limits; otherwise, we would need to know or estimate features to supplement transitions, e.g., \citet{kohjima_learning_2020}'s setting, where the number $k$ of such transitions is missing. 

\subsection{Algorithm: Stein Variational Gradient Descent (SVGD) for Thompson Sampling in Restless Bandits (RB-Stein)}

To address challenges $(1)$ to $(3)$ (see the beginning of Section \ref{sec:method}), we propose an algorithm that combines our MDP extension of \citet{kohjima_learning_2020}'s likelihood with a novel application of finite-particle SVGD to the RMAB setting. To our knowledge, our approach is the first integration of any \textit{valid} finite particle SVGD procedure in the RMAB or RL literature. As we show in Section \ref{sec:studies}, Appendix \ref{supp:trans_sens_kStep}, and Appendix \ref{supp:trans_sens_eta}, leveraging finite particle (mirror) SVGD affords our algorithm computational efficiency and high-fidelity transition probability estimation under demanding conditions. Next, we briefly describe SVGD.

\begin{figure}
    \centering
    \includegraphics[width=.7\linewidth, height =.4\linewidth]{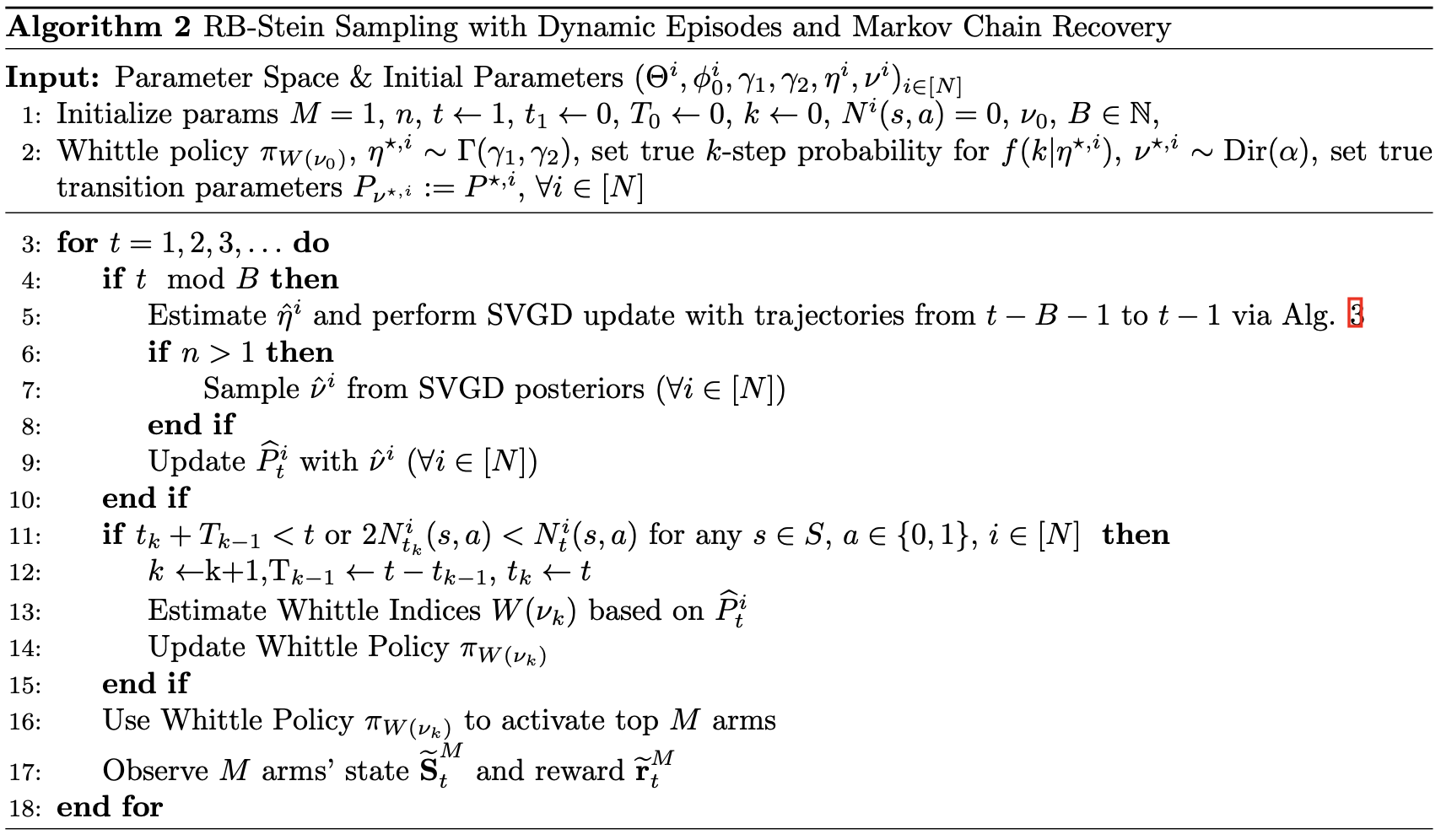}
    \caption{TSDE version of RB-Stein. $N^{i}_{t}(s,a)$ is the number of times arm $i$ has visited $(s,a)$ up to time $t$. $\Tilde{\State}^{M}_{t}$ and $\tilde{r}^{M}_{t}$ are the state and reward vectors of the top $M$ arms observed at time $t$.}
    \label{alg:alg2}
\end{figure}

\begin{figure}
    \centering
    \includegraphics[width=.7\linewidth, height =.23\linewidth]{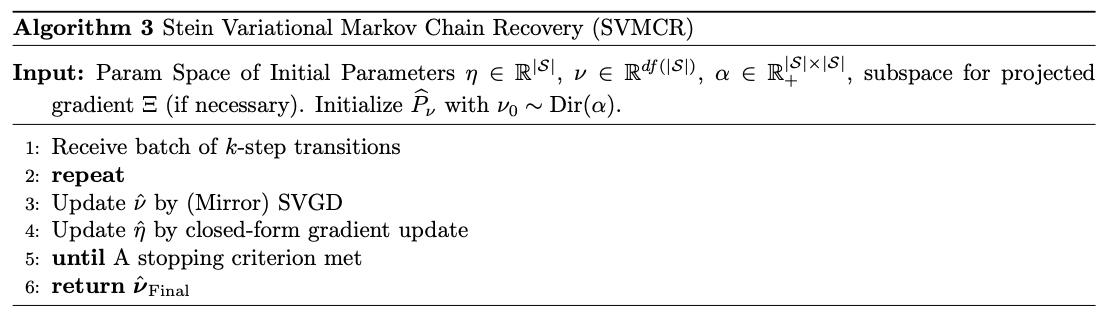}
    \caption{Transition Estimation Algorithm Description. $df(\vert S \vert) := \vert \State \vert \times (\vert \State \vert -1)$. }
    \label{alg:alg3}
\end{figure}

\emph{Posterior Sampling for Partially Observed and Unlabeled Transitions with SVGD.} 
Instead of sampling from the likelihood via Monte Carlo methods or directly estimating parameters by a MM algorithm, e.g., \citet{kohjima_learning_2020}, we use a finite-particle mirror descent implementation of SVGD partially based on \citet{shi_sampling_2022} to obtain updates to our parameters. SVGD is a variational inference (VI) method for generating a deterministic sequence of parameter values (referred to as \textit{particles}) to approximate a target distribution. SVGD is equivalent to iteratively updating each particle in its empirical distribution by taking the steepest minimizing step with respect to the KL divergence (KLD) \citep{liu_stein_2017, liu_stein_2019}. (We refer the reader to Appendix \ref{app:svgd} for a brief review of elementary SVGD updates.) 

SVGD is an ideal method for estimating and sampling posterior transition probability values in the setting of social good applications, as we want to find the best parameter values for the distribution as fast as possible, with minimal exploration (due to ethical concerns) and minimal computational cost. SVGD achieves this in two ways. First, researchers can take advantage of a tunable step-size schedule and a user-defined number of particles for approximating the target distribution. Second, SVGD requires neither estimation of normalizing constants nor knowledge of the true underlying parameters to be evaluated, all while being conveniently parallelizable. We discuss the former point and then the latter.

One way to control the degree of exploration is tunable by choosing a step-size schedule for each SVGD update by working with single-particle vs. $n$-particle SVGD. Another is by using an optimizer that iteratively re-weights the step-size scheduler, e.g., variants of RMSprop or Adam \citep{zhang_improved_2018}. In our work, we work with single-particle SVGD and let RMSprop handle the step-size schedule rather than choosing a specific parameterization for the step-size schedule. However, this leaves a question open about the trade-off between applying single- vs. $n$-particle SVGD.    

Next, single chain gradient descent (or ascent) for MAP is equivalent to single-particle SVGD and has been shown to be very successful in practice, so it follows that even single particle SVGD can generalize well in supervised learning tasks \citep{liu_stein_2019}. Hence, in addition to computational benefits, SVGD allows greater flexibility than typical Monte Carlo methods by allowing researchers to choose particle size \textit{without} serious concerns for generalizability. We test this by strictly working with the single-particle variant in our studies in Section \ref{expts:partner_dyn_est} and in our sensitivity analysis in Appendix \ref{supp:trans_sens_kStep} and \ref{supp:trans_sens_eta}.

\emph{Computational Tractability.} Our method is computationally easier to verify when compared to current standards, e.g., confidence sets of \citet{wang_optimistic_2022}, and our algorithm enjoys several benefits that make it more computationally efficient and deployable to the field. First, being able to choose the number of particles without losing generalizability without needing to estimate normalizing constants for each arm's instance of (M)SVGD updates solves a common bottleneck for upper-confidence bound methods for RMABs. In particular, unlike RMABs relying on upper-confidence bound methods, (M)SVGD does \textit{not} require access to the true transition parameters for its approximation to the target distribution (or target parameters in the single-particle case) to be evaluated \citep{liu_stein_2017, liu_stein_2019, wang_optimal_2022, shi_sampling_2022, shi_finite-particle_2023, das_provably_2023}. This is because optimal kernelized Stein discrepancy (KSD) is not a function of $P^{\star}$; the KSD is $0$ if and only if $\hat{P} = P^{\star}$ \citep{liu_stein_2017, liu_stein_2019, shi_sampling_2022, shi_finite-particle_2023, das_provably_2023}. Finally, recent theoretical results validate that finite-particle KSD is driven to $0$ quickly, i.e., at an order sub-exponential rate \citep{das_provably_2023}, so (M)SVGD provides a fast and accurate approximation to the underlying target distribution (or target parameters in the single-particle case) that is easier to evaluate. 

Second, SVGD relies on specifying a Mercer kernel $k_x$ associated with a Reproducing Kernel Hilbert Space (RKHS), and the choice of $k_x$ can control the convergence properties of SVGD. For example, setting $k_x$ to be the inverse multiquadric kernel yields convergence control properties depending on the choice of its precision parameter \citep{gorham_measuring_2015, shi_sampling_2022}. Adjusting $k_x$ also allows researchers to exploit different forms of SVGD as well as any known structure in the target density $p$. For example, this can improve parameter estimation near boundaries of the target density's parameter space, e.g., when log-concavity in $p$ exists \citep{shi_sampling_2022}.

Third, estimating SVGD per-arm is massively parallelizable. This is because computing each instance of SVGD only requires updating empirical means. In the RMAB setting, where we may have as many as $N$ separate posterior distributions maintain and update, this endows SVGD with a significant reduction in computational cost over typical Monte Carlo-based methods that require maintenance and averaging over many points. Hence, even if each arm maintains its own posterior, it is orders of magnitude cheaper than competitors when properly parallelized. Finally, considering the $n$-particle case, the biggest computational bottleneck would come from computing the score function for all of the $n$ particles. In response, one can easily approximate the score function by sub-sampling mini-batches of the $n$ points and using an appropriate weighted average of the likelihood (see \citet{liu_stein_2019} for details). 

\emph{Valid Estimation.} Finally, we introduce the first \textit{valid} and practical integration of SVGD into model-based RL, which may be of independent interest outside of RMAB applications. To the best of our knowledge, ours is the first algorithm in the RMAB literature that incorporates SVGD. Specifically, each arm's SVGD instance is constrained to estimating parameters within the unit simplex. This motivates two natural changes to SVGD: (i) a simple projection of SVGD particles (PSVGD) onto the probability simplex and (ii) an adjusted mirror-descent variant of SVGD (MSVGD) that automatically respects simplex constraints. While (i) is simpler to implement, it produces volatile estimates when the true parameter is near the boundary of the parameter space \citep{shi_sampling_2022}.

In general, the closest work to ours is in single-agent RL, by \citet{chakraborty_posterior_2022}. \citet{chakraborty_posterior_2022} incorporate classic SVGD into a single agent RL algorithm that also relies on Thompson sampling but their implementation differs from ours in three main respects. First, they rely on an infinite-particle version of SVGD, whereas our algorithm relies on more realistic finite-particle SVGD. Consequently, although they offer theoretical regret guarantees, their guarantees do not quantify the worse-case performance of the algorithm that one would actually implement in practice. Additionally, their infinite-particle SVGD relies on a Bayesian core set construction that lends extra problem structure, whereas we do not. Second, their implementation of SVGD does not enforce simplex constraints, so that transition parameters may violate the support of the target distribution. It is also possible, as a result, that their procedure may yield volatile estimates near boundaries of the parameter space. In contrast, we apply MSVGD to respect simplex constraints while estimating transition probabilities. Third, a direct application of \citet{chakraborty_posterior_2022}'s method would be impossible to execute in practice due to the RMAB's combinatorial state space. So we cannot use it for RMAB applications.

\section{Results}\label{sec:studies}

We compare our proposed algorithm to other common RMAB baselines on Leket and BG data. Sensitivity analyses to $k$-step trajectory data independent of our full RMAB algorithm are provided in Appendix \ref{supp:trans_sens_kStep} and \ref{supp:trans_sens_eta}.

\subsection{RMABs with Controlled Restarts}\label{data:design}

The dynamics model we used to simulate the Leket and BG decision environments is referred to as the RMABs with Controlled Restarts Environment (CRE). The CRE describes any scenario where, upon acting on the target system, it resets to some default state and a different set of transition dynamics. This effectively means that in the CRE setting, our RMAB's problem is to discover how to optimally reset $M$ out of $N$ Markov processes at each time step. While the CRE can be customized to better fit both Leket's and BG's environments, it captures the most salient features across both problems while allowing us to test our RMAB algorithm in an idealized environment. More concretely, we can map the CRE to Leket by regarding successful food rescue as `resetting' the farm from a fecund to a fallow state; for BG, transitioning someone homeless to the next stage of their application `resets' them to a waiting state. For any CRE, however, we must specify how states relate to rewards and the structure of the transition matrix the RMAB must learn.

\citet{akbarzadeh_restless_2019} demonstrated that the CRE environment is indexable, which guarantees that the WIP is optimal \citep{glazebrook_general_2011}. To solve the RMAB problem in the CRE, however, we note that we need to learn only \textit{one} transition matrix. This is because the CRE sets one transition matrix to be a deterministc `reset' matrix that returns the arm to state $1$ if the active action is taken \citep{akbarzadeh_restless_2019}. Explicitly, for an arm $i \in [N]$, the passive action's transition matrix is

\begin{gather}
P^{i}(0)=
  \begin{bmatrix}
   p & q & \ldots & q \\
   q & p & \ldots & q \\
   \vdots & & \ddots & \vdots  \\ 
   q & q & \ldots & p
   \end{bmatrix}.
\end{gather}

Each transition matrix under the passive action is initialized with $p \sim \text{Uniform}\left(\frac{1}{\vert \State \vert}, 1\right)$ and $q = \frac{1-p}{|\State|-1}$. This ensures the matrix is stochastic monotone, which is required for \citet{akbarzadeh_restless_2019}'s indexability result in the CRE. 

Each arm's transition matrix under the active action is the so-called reset matrix, given by $P^{i}(1)=
  \begin{bmatrix}
    \boldsymbol{1}_{|\State| \times 1} & \boldsymbol{0}_{|\State| \times |\State|-1}
   \end{bmatrix}.$

Given an action, each transition matrix is associated with a different state-dependent reward function. 

For every arm $i$, the action-dependent reward function is


\begin{equation}
r(s^i, a^i) = \begin{cases} 
0, & a^i =0 \\
(s^i)^{2}, & a^i =1
\end{cases}.
\end{equation}
This parameterization reflects the Leket and BG problem structure. The algorithm only observes states and rewards under active arms, with $k$-step transitions for a random subset of the active arms. 

We first tested our algorithm vs. a random policy or myopic policy in an environment based on the dynamics estimated from our two real-world datasets. More precisely, we use the same myopic policy as \citet{wang_optimal_2022}. Let $\boldsymbol{w}_{\sigma_i}(t)$ denote the sum over conditional probabilities of each state occurring, with $i \in [N]$ and $(\sigma_j)_{j \in [N]}$ denoting a permutation of $[N]$. Let $\geq_{s}$ denote the first-order stochastic dominance of the left argument over the right. Then if $\boldsymbol{w}_{\sigma_1}(t) \geq_{s} \ldots \geq_{s} \boldsymbol{w}_{\sigma_N}(t)$, this myopic policy chooses the top $M$ ``best'' arms to activate by ranking them according to their relative first-order stochastic dominance at each time step, $\{\sigma_1, \ldots, \sigma_M\} \subset \{0,1\}^N$. We note that this environment satisfies technical conditions outlined by \citet{wang_optimal_2022} and should also guarantee optimality of the myopic policy.

For modeling the $k$-step transitions, we work with a Poisson model and obtain the final gradient estimate in closed form by computing $\lambda_i = \log \mu_i$, where we denote an exponential family's mean-value parameter by $\mu$ and its natural parameter by $\eta = \lambda$. \footnote{As shown by \citet{kohjima_learning_2020}, for many exponential models we can update the gradient of $\eta$ in closed form by following the general expression for any gradient of an exponential family distribution. We note that switching between mean-value parameter $\mu$ and natural parameter $\eta$ may sometimes require numerical computation, e.g., Newton's method for the zero-truncated Poisson distribution.}

\paragraph{Controlled Restart Simulation with $k$-Step Trajectory Data.} We outline the steps we take to run simulations in the RCE below. 

\begin{enumerate}
    \item For each arm $i \in [N]$, independently generate a true $P^{\star, i}(0)$. (If using estimated dynamics from real-world data, use those for $P^{\star, i}(0)$.)
    \item Generate true $k$-step transition model with parameters $\lambda^{\star^{i}}$ and initialize with $\lambda^{i}$. (If using estimated dynamics from real-world data, use those for $\lambda^{\star^{i}}$.)
    \item Starting with a randomly instantiated $P^{i}(0)$ for every arm with parameters $\nu^{i}$, update parameters every $B$ time steps by SVGD and estimate $\lambda^{i}$ by an exact gradient update. 
    \item Run algorithm using its estimated $P^{i}(0)$. Select $M$ arms according to policy; observe states and rewards for the $M$ selected arms each timestep.
\end{enumerate}

\subsection{Comparison to Baselines}

We treat each full dataset as one long vector when fitting the transition matrix model. From there, we use the transition matrix model as the true underlying model and evolve a simulation of $T$ time steps. We set $\vert \State \vert = 2$ states for Leket and $\vert \State \vert = 4$ states for BG. After running for $\frac{T}{2}$ time steps as burn-in, we compare the average reward across the random, myopic, and RB-Stein algorithms.

Finally, in the process of testing our algorithm, we initially considered two variants of SVGD that respect simplex constraints: 
MSVGD and a simpler projected SVGD (PSVGD). MSVGD was superior to PSVGD in all cases and always respects simplex constraints by design, so we only use MSVGD in our algorithm.

\subsection{Estimating Dynamics from Real-World Data}\label{expts:partner_dyn_est} 
Leket's dataset contains information about $2,556$ farmers from March $2010$ to May $2021$ on a day-by-day basis for a total of $3,616$ unique interactions across $6,050$ farmers logged. For each date and farmer, we assign a state of $0:$ \textit{produce unavailable} and a state of $1:$ \textit{produce available}. To handle multiple interactions on the same date, after consultation with Leket we took the max of the states $\{0,1\}$ if there are multiple states logged for a farmer on a single date. From there, we assigned $-1$ to indicate a transition without state information for any date not containing $0$ or $1$ in each farmer's $3,616$-length trajectory. Covariate data provided by Leket was too sparse to be useful. Based on subject-matter considerations, we assume that each farmer is statistically identical and fitted a single transition matrix based on $3$ different levels of sparsity in trajectory data using our algorithm. The first was the top $9$ farmers with at least $1,290$ active states, the second was the top $143$ with at least $3,445$ active states, and the third was the top $545$ farmers with over $5,000$ active states.

Breaking Ground's dataset contains information on $5,904$ unique clients experiencing homelessness from October 2, 2007 to June 8, 2021. Each day, the outreach team canvases neighborhoods in New York and logs information about every attempted engagement with a client, totalling $776,233$ case notes. From each engagement, we extract a state of 0: last known state of client to be on the streets; a state of 1: in stabilized housing (e.g., homeless shelters); a state of 2: in transitional housing (e.g., church beds), or a state of 3: in permanent housing, with the most favorable state being 3. On average, clients had 75 months of engagement with outreach workers.

For data privacy concerns, we converted the client-by-features dataset $\tilde{\boldsymbol{X}} \in \mathbb{R}^{c \times f}$ into a collection of client-by-time trajectories $\textbf{X} \in \mathbb{R}^{c \times t}$. Once obtained, we fit a $k$-step transition model and set it as the true underlying model for generating realizations of $k$, denoted $\hat{\eta}^{\star}$ for $f(\cdot \vert \eta)$. Similarly, we set the true underlying transition parameters for each arm as $\hat{\nu}^{\star}$ based on Algorithm \ref{alg:alg3}. Together, these dictate the unknown dynamics for the RMAB simulation for each arm.

\subsection{Single Particle Performance with Mirror SVGD}\label{single_particle:msvgd}

We present results for our algorithm fit on the Leket and BG data compared to two standard baselines: a random policy and a myopic policy. After fitting the true, unknown parameters (see Section \ref{expts:partner_dyn_est}), we run the transition estimation algorithm in batches of size $B$. We choose to work with the WIP as RB-Stein's deterministic policy for all experiments, as it is still treated as a gold standard in the literature \citep{mate_field_2021, akbarzadeh_learning_2022, wang_optimal_2022}. For our policy update criterion, for this environment we adopt the Thompson Sampling with Dynamic Episodes (TSDE) criterion, as it has been shown to work well for policy updates in this environment \citep{akbarzadeh_learning_2022, jung_thompson_2019}.

We then study and compare each policy's performance in terms of cumulative reward after a burn-in period of $\frac{T}{2}$ time steps. We write RB-Stein with TSDE's update rule as \textit{TS with Markov Chain Recovery (MCR)} in the plots below.

\emph{Leket.} 
In terms of cumulative reward, both the myopic and random policies under-perform significantly with respect to RB-Stein with WIP as the deterministic policy. Error bars denote the estimated standard errors across $N_p$ independent simulation runs and offer evidence that these differences are detectably different from one-another. \\ \newline
\emph{Breaking Ground.}
As with the Leket data, BG exhibits the same rank-ordering of cumulative reward from highest to lowest: TS MCR, Myopic, and then Random.

\begin{figure}[!ht]
 \centering
     \includegraphics[width=.33\textwidth]{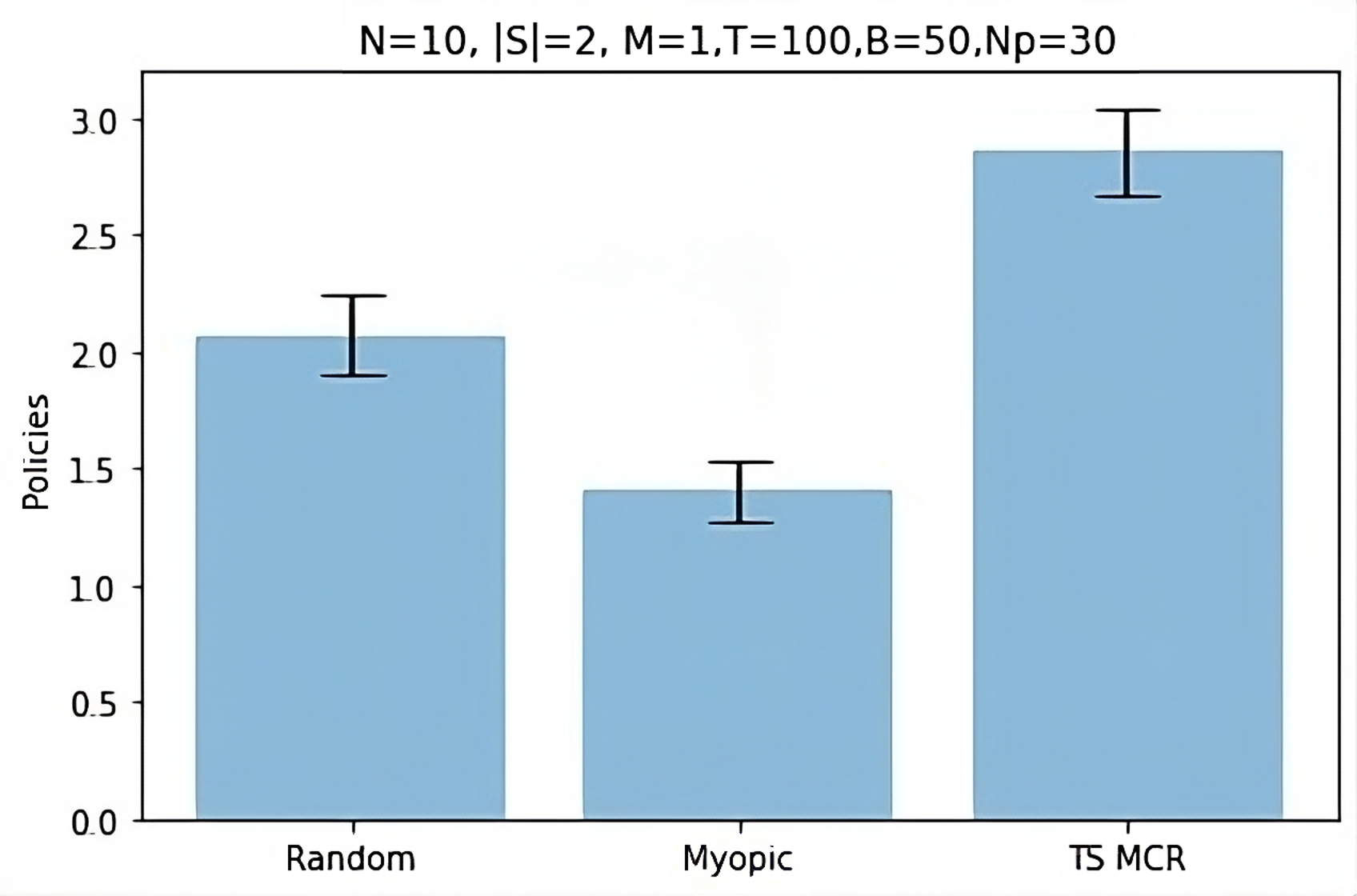}
     \includegraphics[width=.33\textwidth]{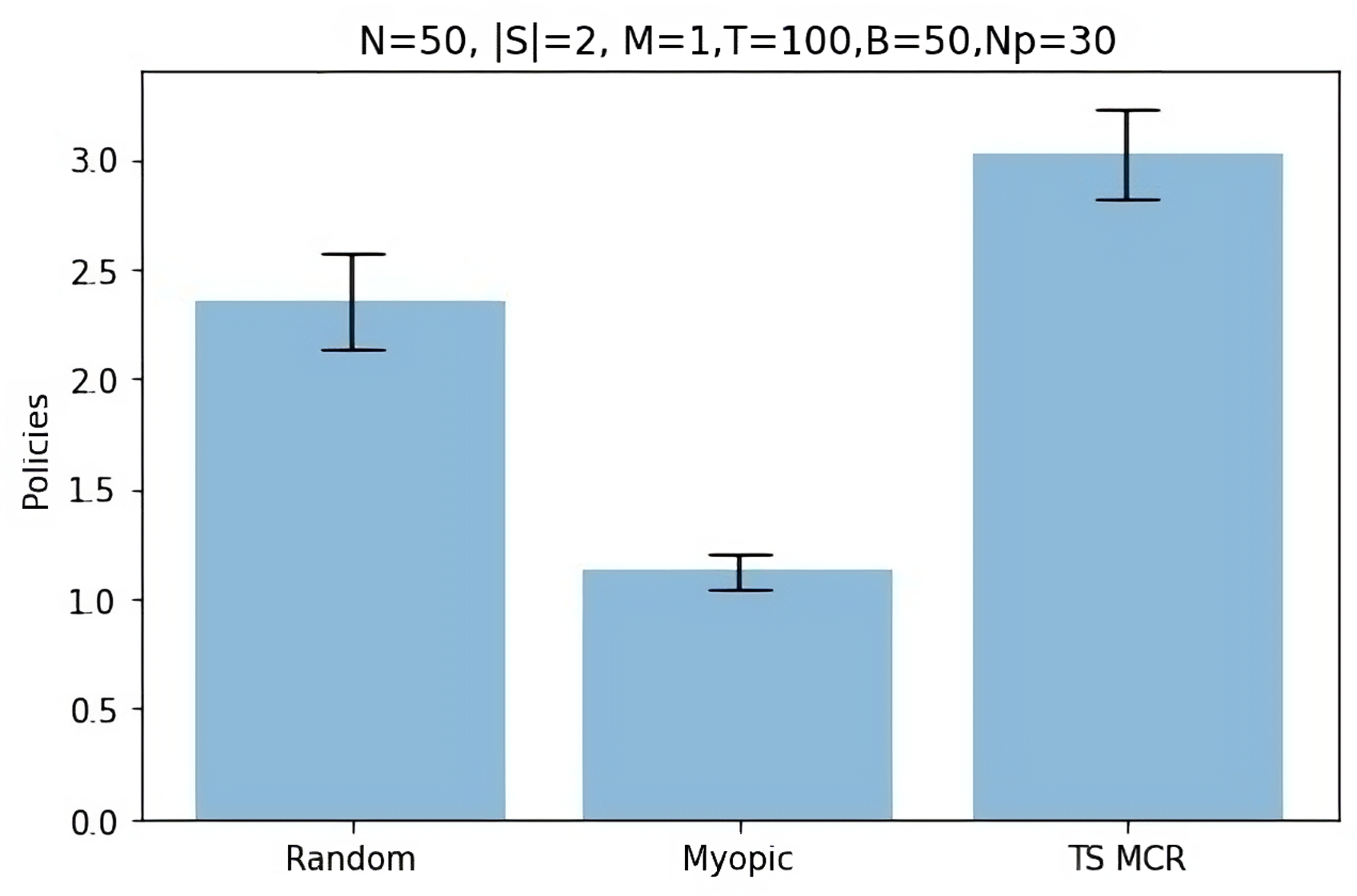}
     \includegraphics[width=.33\textwidth]{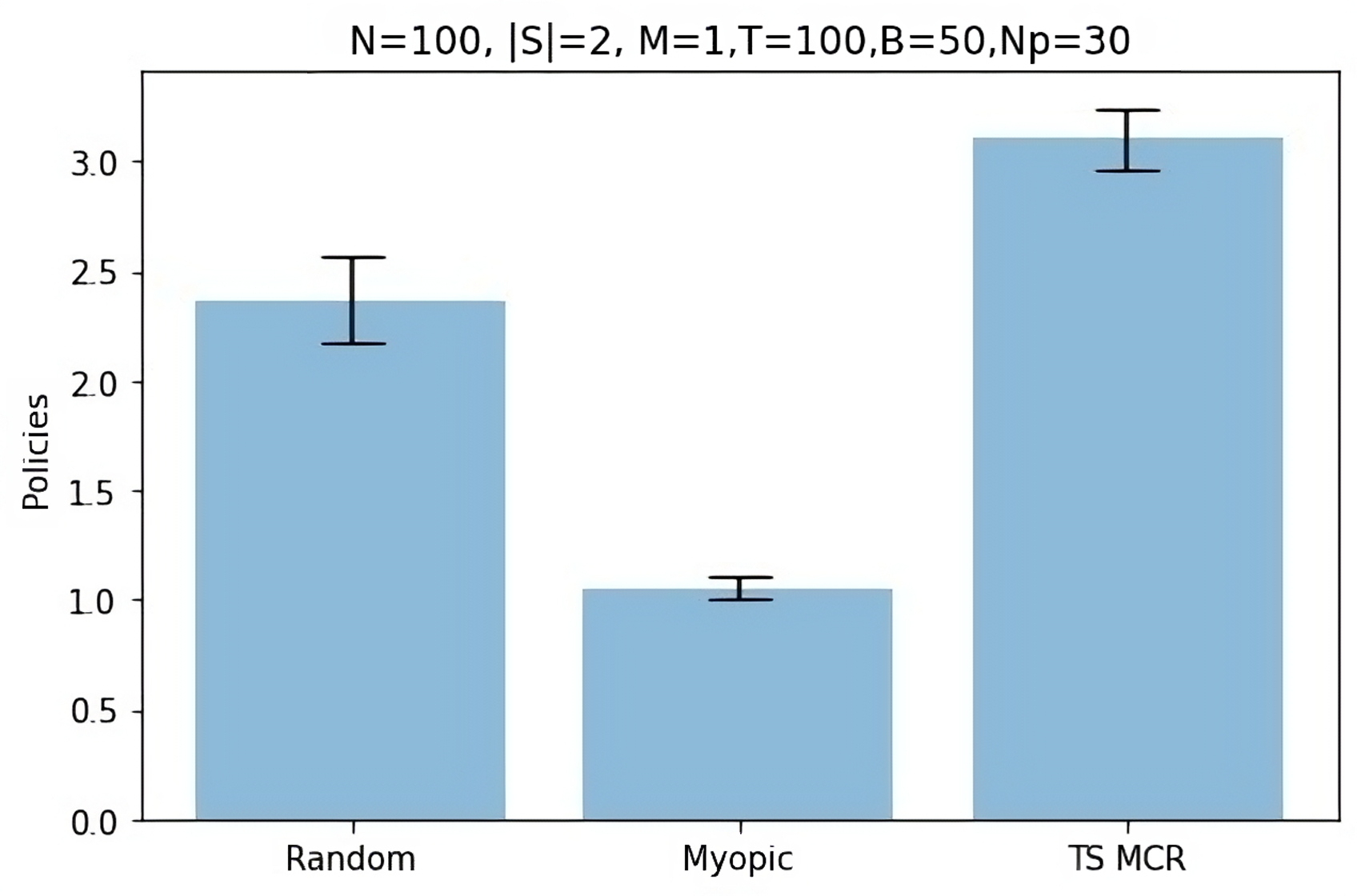}

    \caption{RB-Stein with Mirror SVGD and Dynamic Episodes (TS MCR) : Leket. From left to right: $N = 10$, $N=50$, and $N=100$ arms.}
    \label{fig:leket}
\end{figure}

\begin{figure}[!ht]
     \centering
         \centering
         \includegraphics[width=.33\textwidth]{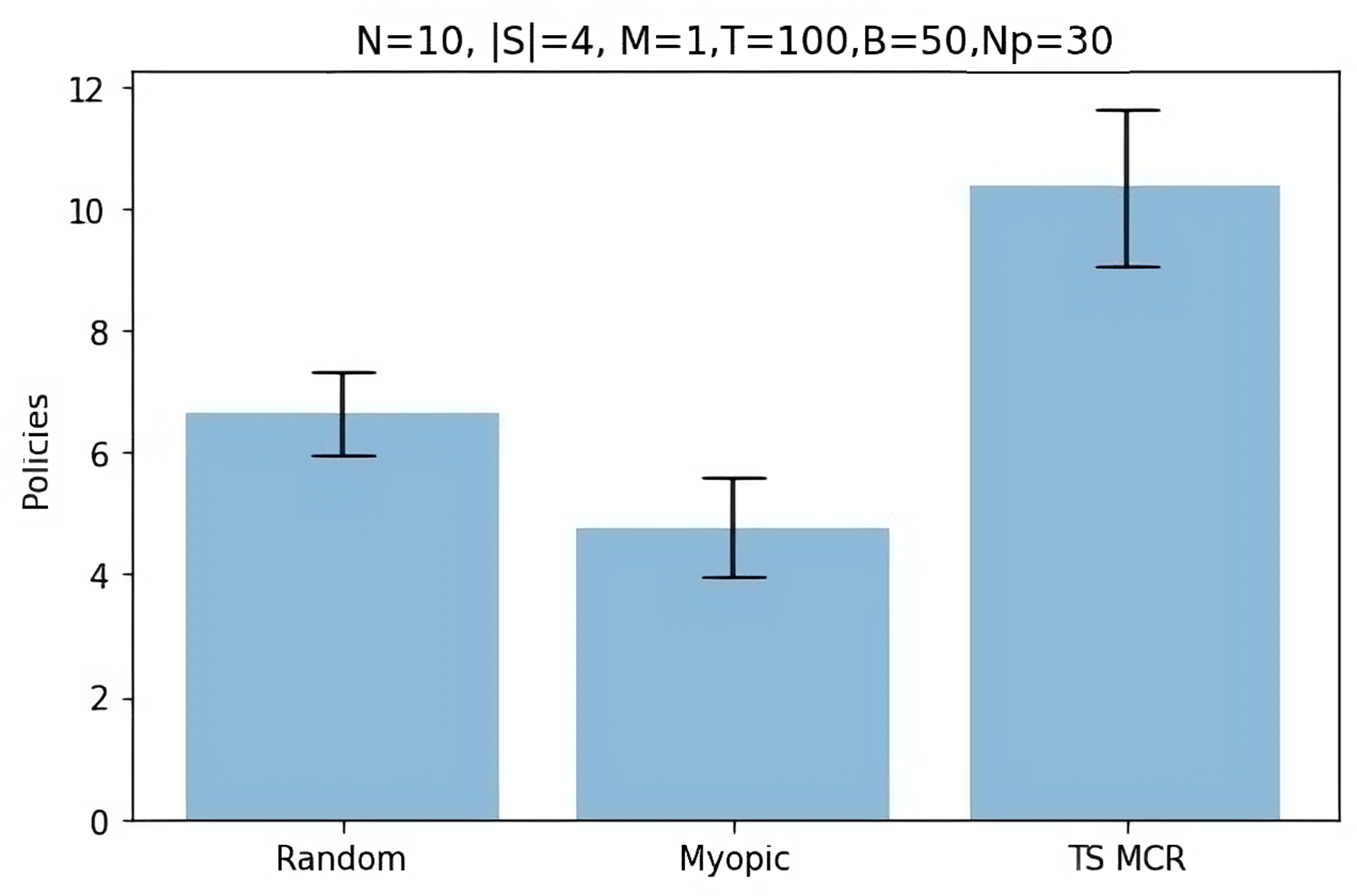}
         \includegraphics[width=.33\textwidth]{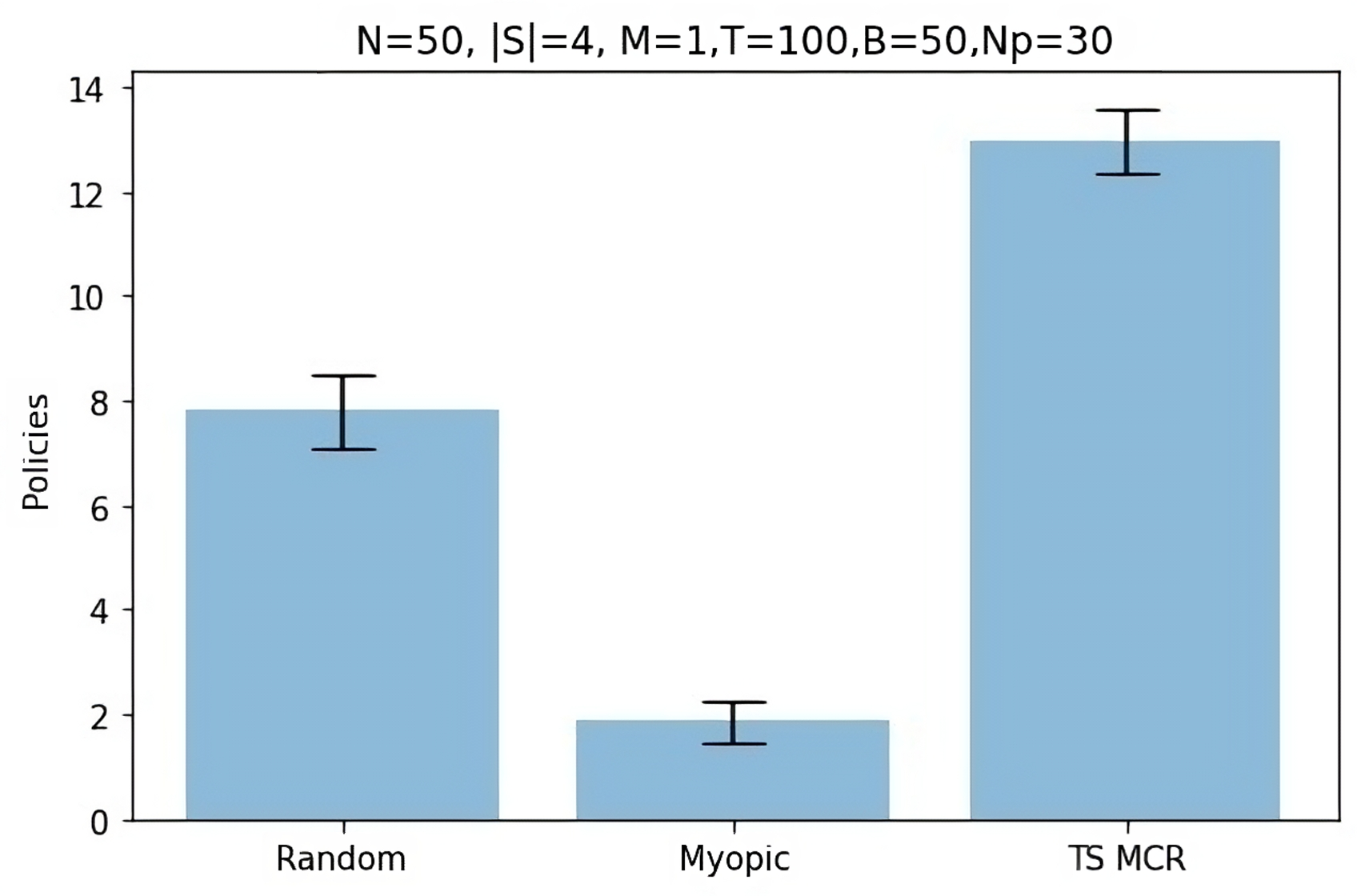}
         \includegraphics[width=.33\textwidth]{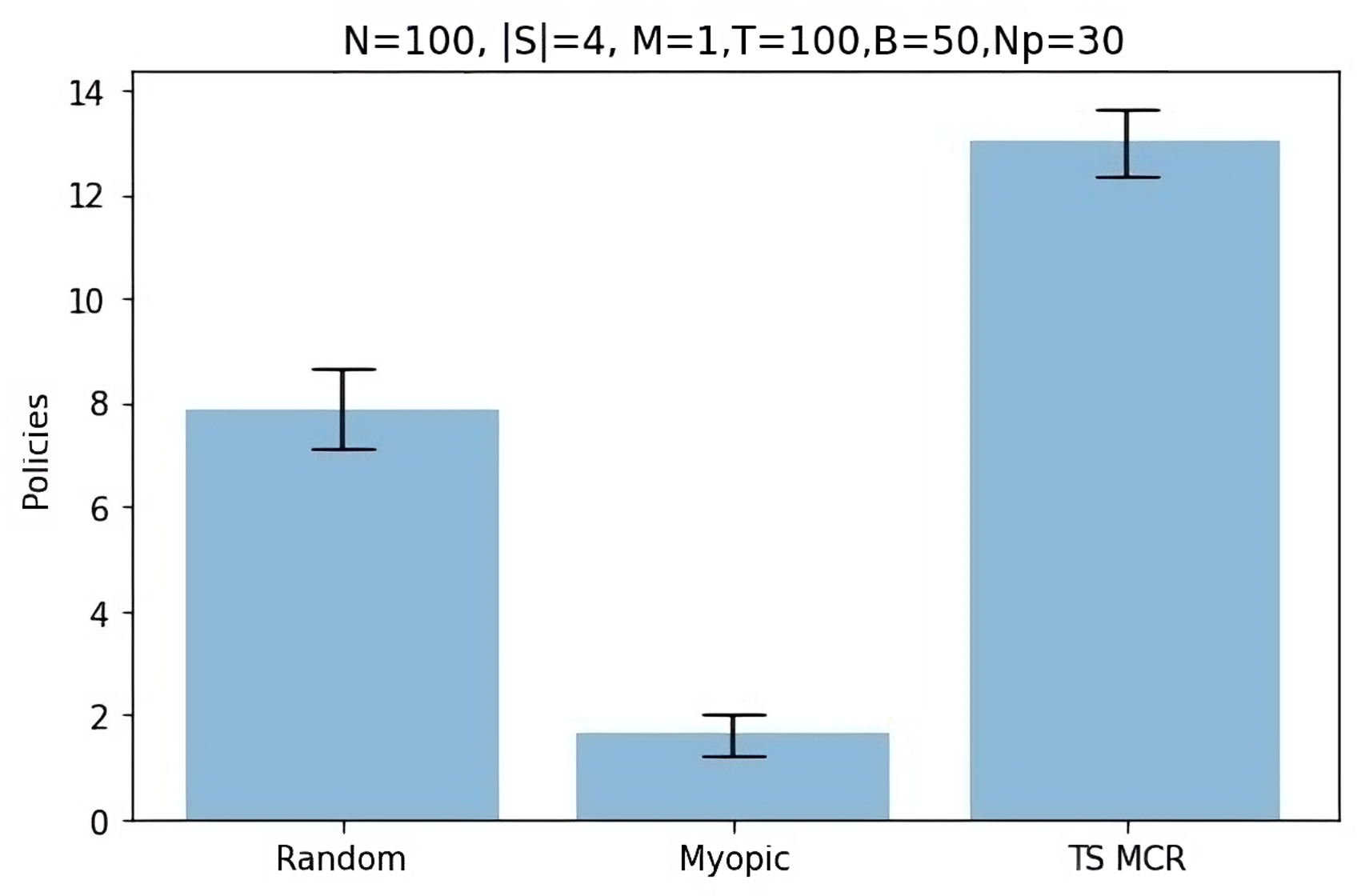}

    \caption{RB-Stein with Mirror SVGD and Dynamic Episdoes (TS MCR): Breaking Ground. From left to right: $N = 10$, $N=50$, and $N=100$ arms.}
    \label{fig:bg}
\end{figure}

\hspace{3mm}
\subsection{Cumulative Change in Reward}

After showing the efficacy of our instance of RB-Stein in the previous section on Leket and BG data, we consider a way to further compare our model's synthetic performance to the real-world data. Ideally, we would run our model on a subset of farms or clients and compare its performance to human counterparts. For practical reasons, we were unable to run this study, which prompted the following. 

For exposition, we chose to work with Leket's data because it had fewer unlabeled state transitions than BG.  When transitions were unlabeled, we assigned the highest-possible reward. In effect, our estimate of the average (CRE) reward is an upward-biased estimate of the rewards. This allowed us to compare the average reward across all clients to the average reward accrued by our RB-Stein instance over the same number of time steps. Next we explain how we fit our RMAB model and compared it to Leket's data. 

First, we split the Leket data in half ($\frac{T}{2}$ time-steps each). Then we fit our RB-Stein instance on the first $\frac{T}{2}$ half of the Leket data. We then record the RMAB's performance on the remaining $\frac{T}{2}$ time-steps. Finally, we compared the RMAB instance's cumulative average reward across arms to the average upward-biased rewards of Leket's farmers over the same $\frac{T}{2}$ period. 

\begin{figure}
     \centering
         \centering
         \includegraphics[width=.425\linewidth, height =.325\linewidth]{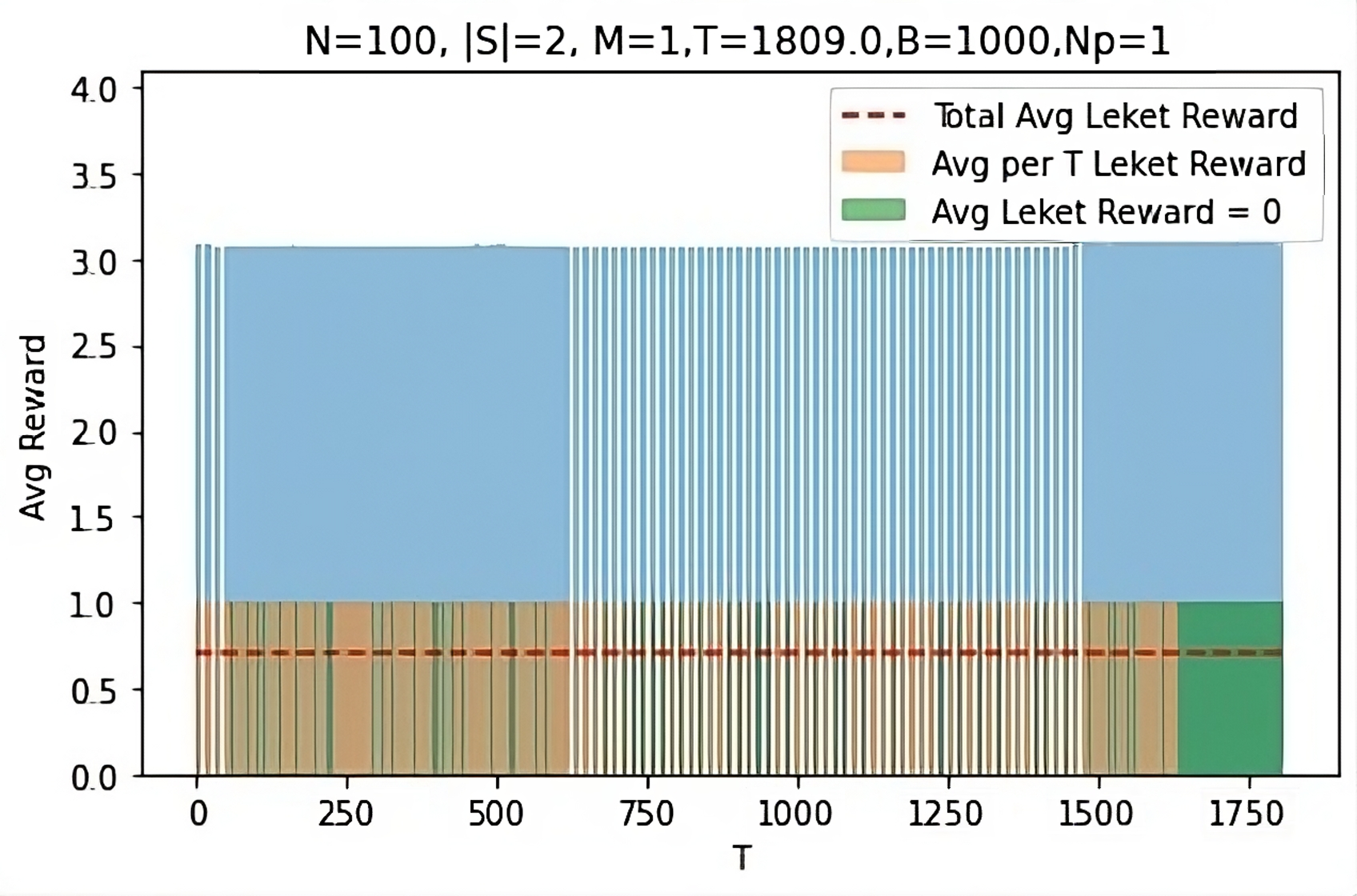}
    \caption{Reward Performance of Proposed Algorthim vs. Observed Rewards from Leket}
    \label{fig:cum_change}
\end{figure}

In Figure \ref{fig:cum_change}, the orange bars represent the average reward \textit{across all farmers} at that time point while the light blue bars show the average reward across the RMAB's arms. We note with green bars when the average Leket reward across farmers was $0$. Comparing the light blue bars to the orange bars, the RMAB produced roughly triple the average reward of the observed states in Leket's data. To contextualize this result, first we recall the CRE's reward structure. The best reward at any time step is when the RMAB activates an arm that is in state $2$, yielding a reward of $4$. When an arm is activated in state $1$, the RMAB earns a reward of $1$, and otherwise earns nothing under the passive action. Our RMAB only considers selecting $M=1$ arm, so the best possible reward for this study is if the RMAB activates an arm in state $2$ every time step. If the RMAB's policy randomly selected an arm each time step, it would receive an average reward of $2.5$. Therefore, if the RMAB's average reward tends to be higher than $2.5$ on average, it learned a policy that earned the optimal reward more often than not. Looking at the plot, we can see that this is the case, as the average reward for the RMAB hovers around $3$. Meanwhile, the average Leket reward hovers around $1$. 

\section{Discussion}

In this paper, we described our work on a common reusable, tractable, and performant transition estimation and RMAB algorithm for two disparate social good applications: gleaning food from farms in Israel by Leket and transitioning homeless clients in New York to permanent housing by Breaking Ground. 

On the technical side, we proposed a TS algorithm to accommodate finite-state RMABs under partial observation and unlabeled ($k$-step) state transitions. In relation to the RMAB literature, our method is the first RMAB algorithm that accommodates general deterministic policies, partial observation of arms with two or more than two states, and unlabeled transition data in a computationally tractable way. There are three components that are key to our methodology: first, we work with a likelihood that uses $k$-step transition data as given. Second, we adapt (M)SVGD to the RMAB setting for a computationally efficient way to update a large number of RMAB arms. Third, we design our algorithm to allow interchanging between deterministic policies, rather than restricting ourselves to the WIP. 

However, our approach does have important limitations to consider. Unfortunately, we do not offer a comparison to more sophisticated RMAB algorithms, as our algorithm is the first that can be reasonably applied to our intended applications. This could mean that if researchers prefer to work with the WIP from the outset, it is possible that extensions of other methodologies that explicitly rely on the WI's for its guarantees, e.g., \citet{wang_optimal_2022}, may be a more relevant choice. Additionally, while we gain many benefits from applying SVGD to RMABs, our implementation is constrained to exponential family $k$-step dynamics and target distributions whose score functions we can write analytically.

On the side of broader impact, this paper represents the first prospective study showing that a deployable algorithm can work with minimal adjustment for several different social change organizations. Therefore, a fruitful strategic direction for the data science for social good space is to consider more general algorithmic formulations and solutions rather than bespoke ones.

  This work was partially funded by the NHLBI T32 Training Grant. 

\bibliographystyle{ACM-Reference-Format}
\bibliography{references,sample-base}


\begin{thebibliography}{52}


\ifx \showCODEN    \undefined \def \showCODEN     #1{\unskip}     \fi
\ifx \showDOI      \undefined \def \showDOI       #1{#1}\fi
\ifx \showISBNx    \undefined \def \showISBNx     #1{\unskip}     \fi
\ifx \showISBNxiii \undefined \def \showISBNxiii  #1{\unskip}     \fi
\ifx \showISSN     \undefined \def \showISSN      #1{\unskip}     \fi
\ifx \showLCCN     \undefined \def \showLCCN      #1{\unskip}     \fi
\ifx \shownote     \undefined \def \shownote      #1{#1}          \fi
\ifx \showarticletitle \undefined \def \showarticletitle #1{#1}   \fi
\ifx \showURL      \undefined \def \showURL       {\relax}        \fi
\providecommand\bibfield[2]{#2}
\providecommand\bibinfo[2]{#2}
\providecommand\natexlab[1]{#1}
\providecommand\showeprint[2][]{arXiv:#2}

\bibitem[Akbarzadeh and Mahajan(2019)]%
        {akbarzadeh_restless_2019}
\bibfield{author}{\bibinfo{person}{Nima Akbarzadeh} {and} \bibinfo{person}{Aditya Mahajan}.} \bibinfo{year}{2019}\natexlab{}.
\newblock \showarticletitle{Restless bandits with controlled restarts: {Indexability} and computation of {Whittle} index}. In \bibinfo{booktitle}{\emph{2019 {IEEE} 58th {Conference} on {Decision} and {Control} ({CDC})}}. \bibinfo{pages}{7294--7300}.
\newblock
\urldef\tempurl%
\url{https://doi.org/10.1109/CDC40024.2019.9029182}
\showDOI{\tempurl}
\newblock
\shownote{ISSN: 2576-2370}.


\bibitem[Akbarzadeh and Mahajan(2022)]%
        {akbarzadeh_learning_2022}
\bibfield{author}{\bibinfo{person}{Nima Akbarzadeh} {and} \bibinfo{person}{Aditya Mahajan}.} \bibinfo{year}{2022}\natexlab{}.
\newblock \bibinfo{title}{On learning {Whittle} index policy for restless bandits with scalable regret}.
\newblock
\newblock
\urldef\tempurl%
\url{http://arxiv.org/abs/2202.03463}
\showURL{%
\tempurl}
\newblock
\shownote{arXiv:2202.03463 [cs, eess]}.


\bibitem[Auer(2003)]%
        {auer_using_2003}
\bibfield{author}{\bibinfo{person}{Peter Auer}.} \bibinfo{year}{2003}\natexlab{}.
\newblock \showarticletitle{Using confidence bounds for exploitation-exploration trade-offs}.
\newblock \bibinfo{journal}{\emph{The Journal of Machine Learning Research}}  \bibinfo{volume}{3} (\bibinfo{date}{March} \bibinfo{year}{2003}), \bibinfo{pages}{397--422}.
\newblock
\showISSN{1532-4435}


\bibitem[Avrachenkov and Borkar(2019)]%
        {avrachenkov_learning_2019}
\bibfield{author}{\bibinfo{person}{Konstantin Avrachenkov} {and} \bibinfo{person}{Vivek~S. Borkar}.} \bibinfo{year}{2019}\natexlab{}.
\newblock \showarticletitle{A learning algorithm for the {Whittle} index policy for scheduling web crawlers}. In \bibinfo{booktitle}{\emph{2019 57th {Annual} {Allerton} {Conference} on {Communication}, {Control}, and {Computing} ({Allerton})}}. \bibinfo{pages}{1001--1006}.
\newblock
\urldef\tempurl%
\url{https://doi.org/10.1109/ALLERTON.2019.8919743}
\showDOI{\tempurl}


\bibitem[Avrachenkov and Borkar(2021)]%
        {avrachenkov_whittle_2021}
\bibfield{author}{\bibinfo{person}{Konstantin~E. Avrachenkov} {and} \bibinfo{person}{Vivek~S. Borkar}.} \bibinfo{year}{2021}\natexlab{}.
\newblock \showarticletitle{Whittle index based {Q}-learning for restless bandits with average reward}.
\newblock \bibinfo{journal}{\emph{arXiv:2004.14427 [cs, math, stat]}} (\bibinfo{date}{Sept.} \bibinfo{year}{2021}).
\newblock
\urldef\tempurl%
\url{http://arxiv.org/abs/2004.14427}
\showURL{%
\tempurl}
\newblock
\shownote{arXiv: 2004.14427}.


\bibitem[Ayer et~al\mbox{.}(2019)]%
        {ayer_prioritizing_2019}
\bibfield{author}{\bibinfo{person}{Turgay Ayer}, \bibinfo{person}{Can Zhang}, \bibinfo{person}{Anthony Bonifonte}, \bibinfo{person}{Anne~C. Spaulding}, {and} \bibinfo{person}{Jagpreet Chhatwal}.} \bibinfo{year}{2019}\natexlab{}.
\newblock \showarticletitle{Prioritizing {Hepatitis} {C} {Treatment} in {U}.{S}. {Prisons}}.
\newblock \bibinfo{journal}{\emph{Operations Research}} \bibinfo{volume}{67}, \bibinfo{number}{3} (\bibinfo{date}{May} \bibinfo{year}{2019}), \bibinfo{pages}{853--873}.
\newblock
\showISSN{0030-364X}
\urldef\tempurl%
\url{https://doi.org/10.1287/opre.2018.1812}
\showDOI{\tempurl}
\newblock
\shownote{Publisher: INFORMS}.


\bibitem[Bellamy et~al\mbox{.}(2018)]%
        {bellamy_ai_2018}
\bibfield{author}{\bibinfo{person}{Rachel K.~E. Bellamy}, \bibinfo{person}{Kuntal Dey}, \bibinfo{person}{Michael Hind}, \bibinfo{person}{Samuel~C. Hoffman}, \bibinfo{person}{Stephanie Houde}, \bibinfo{person}{Kalapriya Kannan}, \bibinfo{person}{Pranay Lohia}, \bibinfo{person}{Jacquelyn Martino}, \bibinfo{person}{Sameep Mehta}, \bibinfo{person}{Aleksandra Mojsilovic}, \bibinfo{person}{Seema Nagar}, \bibinfo{person}{Karthikeyan~Natesan Ramamurthy}, \bibinfo{person}{John Richards}, \bibinfo{person}{Diptikalyan Saha}, \bibinfo{person}{Prasanna Sattigeri}, \bibinfo{person}{Moninder Singh}, \bibinfo{person}{Kush~R. Varshney}, {and} \bibinfo{person}{Yunfeng Zhang}.} \bibinfo{year}{2018}\natexlab{}.
\newblock \bibinfo{title}{{AI} {Fairness} 360: {An} {Extensible} {Toolkit} for {Detecting}, {Understanding}, and {Mitigating} {Unwanted} {Algorithmic} {Bias}}.
\newblock
\newblock
\urldef\tempurl%
\url{http://arxiv.org/abs/1810.01943}
\showURL{%
\tempurl}
\newblock
\shownote{arXiv:1810.01943 [cs]}.


\bibitem[Borkar and Chadha(2018)]%
        {borkar_reinforcement_2018}
\bibfield{author}{\bibinfo{person}{Vivek~S. Borkar} {and} \bibinfo{person}{Karan Chadha}.} \bibinfo{year}{2018}\natexlab{}.
\newblock \showarticletitle{A reinforcement learning algorithm for restless bandits}. In \bibinfo{booktitle}{\emph{2018 {Indian} {Control} {Conference} ({ICC})}}. \bibinfo{pages}{89--94}.
\newblock
\urldef\tempurl%
\url{https://doi.org/10.1109/INDIANCC.2018.8307959}
\showDOI{\tempurl}


\bibitem[Chakraborty et~al\mbox{.}(2022)]%
        {chakraborty_posterior_2022}
\bibfield{author}{\bibinfo{person}{Souradip Chakraborty}, \bibinfo{person}{Amrit~Singh Bedi}, \bibinfo{person}{Alec Koppel}, \bibinfo{person}{Brian~M. Sadler}, \bibinfo{person}{Furong Huang}, \bibinfo{person}{Pratap Tokekar}, {and} \bibinfo{person}{Dinesh Manocha}.} \bibinfo{year}{2022}\natexlab{}.
\newblock \bibinfo{title}{Posterior {Coreset} {Construction} with {Kernelized} {Stein} {Discrepancy} for {Model}-{Based} {Reinforcement} {Learning}}.
\newblock
\newblock
\urldef\tempurl%
\url{http://arxiv.org/abs/2206.01162}
\showURL{%
\tempurl}
\newblock
\shownote{arXiv:2206.01162 [cs, math, stat]}.


\bibitem[Das and Nagaraj(2023)]%
        {das_provably_2023}
\bibfield{author}{\bibinfo{person}{Aniket Das} {and} \bibinfo{person}{Dheeraj Nagaraj}.} \bibinfo{year}{2023}\natexlab{}.
\newblock \bibinfo{title}{Provably {Fast} {Finite} {Particle} {Variants} of {SVGD} via {Virtual} {Particle} {Stochastic} {Approximation}}.
\newblock
\newblock
\urldef\tempurl%
\url{http://arxiv.org/abs/2305.17558}
\showURL{%
\tempurl}
\newblock
\shownote{arXiv:2305.17558 [cs, math, stat]}.


\bibitem[Durrett(2019)]%
        {durrett_probability_2019}
\bibfield{author}{\bibinfo{person}{Rick Durrett}.} \bibinfo{year}{2019}\natexlab{}.
\newblock \bibinfo{booktitle}{\emph{Probability: {Theory} and {Examples}}}. Vol.~\bibinfo{volume}{49}.
\newblock \bibinfo{publisher}{Cambridge University Press}.
\newblock


\bibitem[Gittins(1979)]%
        {gittins_bandit_1979}
\bibfield{author}{\bibinfo{person}{J.~C. Gittins}.} \bibinfo{year}{1979}\natexlab{}.
\newblock \showarticletitle{Bandit {Processes} and {Dynamic} {Allocation} {Indices}}.
\newblock \bibinfo{journal}{\emph{Journal of the Royal Statistical Society: Series B (Methodological)}} \bibinfo{volume}{41}, \bibinfo{number}{2} (\bibinfo{year}{1979}), \bibinfo{pages}{148--164}.
\newblock
\showISSN{2517-6161}
\urldef\tempurl%
\url{https://doi.org/10.1111/j.2517-6161.1979.tb01068.x}
\showDOI{\tempurl}


\bibitem[Glazebrook et~al\mbox{.}(2011)]%
        {glazebrook_general_2011}
\bibfield{author}{\bibinfo{person}{Kevin~D. Glazebrook}, \bibinfo{person}{David~J. Hodge}, {and} \bibinfo{person}{Chris Kirkbride}.} \bibinfo{year}{2011}\natexlab{}.
\newblock \showarticletitle{{GENERAL} {NOTIONS} {OF} {INDEXABILITY} {FOR} {QUEUEING} {CONTROL} {AND} {ASSET} {MANAGEMENT}}.
\newblock \bibinfo{journal}{\emph{The Annals of Applied Probability}} \bibinfo{volume}{21}, \bibinfo{number}{3} (\bibinfo{year}{2011}), \bibinfo{pages}{876--907}.
\newblock
\showISSN{1050-5164}
\newblock
\shownote{Publisher: Institute of Mathematical Statistics}.


\bibitem[Gorham and Mackey(2015)]%
        {gorham_measuring_2015}
\bibfield{author}{\bibinfo{person}{Jackson Gorham} {and} \bibinfo{person}{Lester Mackey}.} \bibinfo{year}{2015}\natexlab{}.
\newblock \showarticletitle{Measuring {Sample} {Quality} with {Stein}' s {Method}}. In \bibinfo{booktitle}{\emph{Advances in {Neural} {Information} {Processing} {Systems}}}, Vol.~\bibinfo{volume}{28}. \bibinfo{publisher}{Curran Associates, Inc.}
\newblock


\bibitem[Herlihy et~al\mbox{.}(2022)]%
        {herlihy_planning_2022}
\bibfield{author}{\bibinfo{person}{Christine Herlihy}, \bibinfo{person}{Aviva Prins}, \bibinfo{person}{Aravind Srinivasan}, {and} \bibinfo{person}{John~P. Dickerson}.} \bibinfo{year}{2022}\natexlab{}.
\newblock \bibinfo{title}{Planning to {Fairly} {Allocate}: {Probabilistic} {Fairness} in the {Restless} {Bandit} {Setting}}.
\newblock
\newblock
\urldef\tempurl%
\url{http://arxiv.org/abs/2106.07677}
\showURL{%
\tempurl}
\newblock
\shownote{arXiv:2106.07677 [cs]}.


\bibitem[Jung et~al\mbox{.}(2019)]%
        {jung_thompson_2019}
\bibfield{author}{\bibinfo{person}{Young~Hun Jung}, \bibinfo{person}{Marc Abeille}, {and} \bibinfo{person}{Ambuj Tewari}.} \bibinfo{year}{2019}\natexlab{}.
\newblock \showarticletitle{Thompson {Sampling} in {Non}-{Episodic} {Restless} {Bandits}}.
\newblock \bibinfo{journal}{\emph{arXiv:1910.05654 [cs, stat]}} (\bibinfo{date}{Oct.} \bibinfo{year}{2019}).
\newblock
\urldef\tempurl%
\url{http://arxiv.org/abs/1910.05654}
\showURL{%
\tempurl}
\newblock
\shownote{arXiv: 1910.05654}.


\bibitem[Jung and Tewari(2019)]%
        {jung_regret_2019}
\bibfield{author}{\bibinfo{person}{Young~Hun Jung} {and} \bibinfo{person}{Ambuj Tewari}.} \bibinfo{year}{2019}\natexlab{}.
\newblock \bibinfo{title}{Regret {Bounds} for {Thompson} {Sampling} in {Episodic} {Restless} {Bandit} {Problems}}.
\newblock
\newblock
\urldef\tempurl%
\url{http://arxiv.org/abs/1905.12673}
\showURL{%
\tempurl}
\newblock
\shownote{arXiv:1905.12673 [cs, stat]}.


\bibitem[Killian et~al\mbox{.}(2021)]%
        {killian_robust_2021}
\bibfield{author}{\bibinfo{person}{Jackson~A. Killian}, \bibinfo{person}{Lily Xu}, \bibinfo{person}{Arpita Biswas}, {and} \bibinfo{person}{Milind Tambe}.} \bibinfo{year}{2021}\natexlab{}.
\newblock \showarticletitle{Robust {Restless} {Bandits}: {Tackling} {Interval} {Uncertainty} with {Deep} {Reinforcement} {Learning}}.
\newblock \bibinfo{journal}{\emph{arXiv:2107.01689 [cs]}} (\bibinfo{date}{July} \bibinfo{year}{2021}).
\newblock
\urldef\tempurl%
\url{http://arxiv.org/abs/2107.01689}
\showURL{%
\tempurl}
\newblock
\shownote{arXiv: 2107.01689}.


\bibitem[Kohjima et~al\mbox{.}(2020)]%
        {kohjima_learning_2020}
\bibfield{author}{\bibinfo{person}{Masahiro Kohjima}, \bibinfo{person}{Takeshi Kurashima}, {and} \bibinfo{person}{Hiroyuki Toda}.} \bibinfo{year}{2020}\natexlab{}.
\newblock \showarticletitle{Learning with {Labeled} and {Unlabeled} {Multi}-{Step} {Transition} {Data} for {Recovering} {Markov} {Chain} from {Incomplete} {Transition} {Data}}. In \bibinfo{booktitle}{\emph{Proceedings of the {Twenty}-{Ninth} {International} {Joint} {Conference} on {Artificial} {Intelligence}}}. \bibinfo{publisher}{International Joint Conferences on Artificial Intelligence Organization}, \bibinfo{address}{Yokohama, Japan}, \bibinfo{pages}{2412--2419}.
\newblock
\showISBNx{978-0-9992411-6-5}
\urldef\tempurl%
\url{https://doi.org/10.24963/ijcai.2020/334}
\showDOI{\tempurl}


\bibitem[Lai and Robbins(1985)]%
        {lai_asymptotically_1985}
\bibfield{author}{\bibinfo{person}{T.~L Lai} {and} \bibinfo{person}{Herbert Robbins}.} \bibinfo{year}{1985}\natexlab{}.
\newblock \showarticletitle{Asymptotically efficient adaptive allocation rules}.
\newblock \bibinfo{journal}{\emph{Advances in Applied Mathematics}} \bibinfo{volume}{6}, \bibinfo{number}{1} (\bibinfo{date}{March} \bibinfo{year}{1985}), \bibinfo{pages}{4--22}.
\newblock
\showISSN{0196-8858}
\urldef\tempurl%
\url{https://doi.org/10.1016/0196-8858(85)90002-8}
\showDOI{\tempurl}


\bibitem[Lattimore and Szepesvári(2020)]%
        {lattimore_bandit_2020}
\bibfield{author}{\bibinfo{person}{Tor Lattimore} {and} \bibinfo{person}{Csaba Szepesvári}.} \bibinfo{year}{2020}\natexlab{}.
\newblock \bibinfo{booktitle}{\emph{Bandit {Algorithms}: {Solutions}} (\bibinfo{edition}{1} ed.)}.
\newblock \bibinfo{publisher}{Cambridge University Press}.
\newblock
\showISBNx{978-1-108-57140-1 978-1-108-48682-8}
\urldef\tempurl%
\url{https://doi.org/10.1017/9781108571401}
\showDOI{\tempurl}


\bibitem[Lee et~al\mbox{.}(2019)]%
        {lee_optimal_2019}
\bibfield{author}{\bibinfo{person}{Elliot Lee}, \bibinfo{person}{Mariel~S. Lavieri}, {and} \bibinfo{person}{Michael Volk}.} \bibinfo{year}{2019}\natexlab{}.
\newblock \showarticletitle{Optimal {Screening} for {Hepatocellular} {Carcinoma}: {A} {Restless} {Bandit} {Model}}.
\newblock \bibinfo{journal}{\emph{Manufacturing \& Service Operations Management}} \bibinfo{volume}{21}, \bibinfo{number}{1} (\bibinfo{date}{Jan.} \bibinfo{year}{2019}), \bibinfo{pages}{198--212}.
\newblock
\showISSN{1523-4614}
\urldef\tempurl%
\url{https://doi.org/10.1287/msom.2017.0697}
\showDOI{\tempurl}
\newblock
\shownote{Publisher: INFORMS}.


\bibitem[Liu(2017)]%
        {liu_stein_2017}
\bibfield{author}{\bibinfo{person}{Qiang Liu}.} \bibinfo{year}{2017}\natexlab{}.
\newblock \showarticletitle{Stein {Variational} {Gradient} {Descent} as {Gradient} {Flow}}. In \bibinfo{booktitle}{\emph{Advances in {Neural} {Information} {Processing} {Systems}}}, Vol.~\bibinfo{volume}{30}. \bibinfo{publisher}{Curran Associates, Inc.}, \bibinfo{pages}{1--9}.
\newblock


\bibitem[Liu and Wang(2019)]%
        {liu_stein_2019}
\bibfield{author}{\bibinfo{person}{Qiang Liu} {and} \bibinfo{person}{Dilin Wang}.} \bibinfo{year}{2019}\natexlab{}.
\newblock \bibinfo{title}{Stein {Variational} {Gradient} {Descent}: {A} {General} {Purpose} {Bayesian} {Inference} {Algorithm}}.
\newblock
\newblock
\urldef\tempurl%
\url{http://arxiv.org/abs/1608.04471}
\showURL{%
\tempurl}
\newblock
\shownote{arXiv:1608.04471 [cs, stat]}.


\bibitem[Manne(1960)]%
        {manne_linear_1960}
\bibfield{author}{\bibinfo{person}{Alan~S. Manne}.} \bibinfo{year}{1960}\natexlab{}.
\newblock \showarticletitle{Linear {Programming} and {Sequential} {Decisions}}.
\newblock \bibinfo{journal}{\emph{Management Science}} \bibinfo{volume}{6}, \bibinfo{number}{3} (\bibinfo{year}{1960}), \bibinfo{pages}{259--267}.
\newblock
\showISSN{0025-1909}
\newblock
\shownote{Publisher: INFORMS}.


\bibitem[Marecek(2020)]%
        {marecek_screening_2020}
\bibfield{author}{\bibinfo{person}{Jakub Marecek}.} \bibinfo{year}{2020}\natexlab{}.
\newblock \bibinfo{title}{Screening for an {Infectious} {Disease} as a {Problem} in {Stochastic} {Control}}.
\newblock
\newblock
\urldef\tempurl%
\url{http://arxiv.org/abs/2011.00635}
\showURL{%
\tempurl}
\newblock
\shownote{arXiv:2011.00635 [physics, stat]}.


\bibitem[Mate et~al\mbox{.}({[n.\,d.]})]%
        {mate_collapsing_nodate}
\bibfield{author}{\bibinfo{person}{Aditya Mate}, \bibinfo{person}{Jackson~A Killian}, {and} \bibinfo{person}{Haifeng Xu}.} \bibinfo{year}{[n.\,d.]}\natexlab{}.
\newblock \showarticletitle{Collapsing {Bandits} and {Their} {Application} to {Public} {Health} {Interventions}}.
\newblock  (\bibinfo{year}{[n.\,d.]}), \bibinfo{pages}{22}.
\newblock


\bibitem[Mate et~al\mbox{.}(2021)]%
        {mate_field_2021}
\bibfield{author}{\bibinfo{person}{Aditya Mate}, \bibinfo{person}{Lovish Madaan}, \bibinfo{person}{Aparna Taneja}, \bibinfo{person}{Neha Madhiwalla}, \bibinfo{person}{Shresth Verma}, \bibinfo{person}{Gargi Singh}, \bibinfo{person}{Aparna Hegde}, \bibinfo{person}{Pradeep Varakantham}, {and} \bibinfo{person}{Milind Tambe}.} \bibinfo{year}{2021}\natexlab{}.
\newblock \showarticletitle{Field {Study} in {Deploying} {Restless} {Multi}-{Armed} {Bandits}: {Assisting} {Non}-{Profits} in {Improving} {Maternal} and {Child} {Health}}.
\newblock \bibinfo{journal}{\emph{arXiv:2109.08075 [cs]}} (\bibinfo{date}{Oct.} \bibinfo{year}{2021}).
\newblock
\urldef\tempurl%
\url{http://arxiv.org/abs/2109.08075}
\showURL{%
\tempurl}
\newblock
\shownote{arXiv: 2109.08075}.


\bibitem[Mnih et~al\mbox{.}(2015)]%
        {mnih_human-level_2015}
\bibfield{author}{\bibinfo{person}{Volodymyr Mnih}, \bibinfo{person}{Koray Kavukcuoglu}, \bibinfo{person}{David Silver}, \bibinfo{person}{Andrei~A. Rusu}, \bibinfo{person}{Joel Veness}, \bibinfo{person}{Marc~G. Bellemare}, \bibinfo{person}{Alex Graves}, \bibinfo{person}{Martin Riedmiller}, \bibinfo{person}{Andreas~K. Fidjeland}, \bibinfo{person}{Georg Ostrovski}, \bibinfo{person}{Stig Petersen}, \bibinfo{person}{Charles Beattie}, \bibinfo{person}{Amir Sadik}, \bibinfo{person}{Ioannis Antonoglou}, \bibinfo{person}{Helen King}, \bibinfo{person}{Dharshan Kumaran}, \bibinfo{person}{Daan Wierstra}, \bibinfo{person}{Shane Legg}, {and} \bibinfo{person}{Demis Hassabis}.} \bibinfo{year}{2015}\natexlab{}.
\newblock \showarticletitle{Human-level control through deep reinforcement learning}.
\newblock \bibinfo{journal}{\emph{Nature}} \bibinfo{volume}{518}, \bibinfo{number}{7540} (\bibinfo{date}{Feb.} \bibinfo{year}{2015}), \bibinfo{pages}{529--533}.
\newblock
\showISSN{0028-0836, 1476-4687}
\urldef\tempurl%
\url{https://doi.org/10.1038/nature14236}
\showDOI{\tempurl}


\bibitem[Niño-Mora(2001)]%
        {nino-mora_restless_2001}
\bibfield{author}{\bibinfo{person}{José Niño-Mora}.} \bibinfo{year}{2001}\natexlab{}.
\newblock \showarticletitle{Restless {Bandits}, {Partial} {Conservation} {Laws} and {Indexability}}.
\newblock \bibinfo{journal}{\emph{Advances in Applied Probability}} \bibinfo{volume}{33}, \bibinfo{number}{1} (\bibinfo{year}{2001}), \bibinfo{pages}{76--98}.
\newblock
\showISSN{0001-8678}
\newblock
\shownote{Publisher: Applied Probability Trust}.


\bibitem[Niño-Mora(2023)]%
        {nino-mora_markovian_2023}
\bibfield{author}{\bibinfo{person}{José Niño-Mora}.} \bibinfo{year}{2023}\natexlab{}.
\newblock \showarticletitle{Markovian {Restless} {Bandits} and {Index} {Policies}: {A} {Review}}.
\newblock \bibinfo{journal}{\emph{Mathematics}} \bibinfo{volume}{11}, \bibinfo{number}{7} (\bibinfo{date}{Jan.} \bibinfo{year}{2023}), \bibinfo{pages}{1639}.
\newblock
\showISSN{2227-7390}
\urldef\tempurl%
\url{https://doi.org/10.3390/math11071639}
\showDOI{\tempurl}
\newblock
\shownote{Number: 7 Publisher: Multidisciplinary Digital Publishing Institute}.


\bibitem[Osband and Van~Roy(2017)]%
        {osband_why_2017}
\bibfield{author}{\bibinfo{person}{Ian Osband} {and} \bibinfo{person}{Benjamin Van~Roy}.} \bibinfo{year}{2017}\natexlab{}.
\newblock \bibinfo{title}{Why is {Posterior} {Sampling} {Better} than {Optimism} for {Reinforcement} {Learning}?}
\newblock
\newblock
\urldef\tempurl%
\url{http://arxiv.org/abs/1607.00215}
\showURL{%
\tempurl}
\newblock
\shownote{arXiv:1607.00215 [cs, stat]}.


\bibitem[Ouyang et~al\mbox{.}(2017)]%
        {ouyang_learning_2017}
\bibfield{author}{\bibinfo{person}{Yi Ouyang}, \bibinfo{person}{Mukul Gagrani}, \bibinfo{person}{Ashutosh Nayyar}, {and} \bibinfo{person}{Rahul Jain}.} \bibinfo{year}{2017}\natexlab{}.
\newblock \showarticletitle{Learning {Unknown} {Markov} {Decision} {Processes}: {A} {Thompson} {Sampling} {Approach}}.
\newblock \bibinfo{journal}{\emph{arXiv:1709.04570 [cs]}} (\bibinfo{date}{Sept.} \bibinfo{year}{2017}).
\newblock
\urldef\tempurl%
\url{http://arxiv.org/abs/1709.04570}
\showURL{%
\tempurl}
\newblock
\shownote{arXiv: 1709.04570}.


\bibitem[Papadimitriou and Tsitsiklis(1994)]%
        {papadimitriou_complexity_1994}
\bibfield{author}{\bibinfo{person}{C.H. Papadimitriou} {and} \bibinfo{person}{J.N. Tsitsiklis}.} \bibinfo{year}{1994}\natexlab{}.
\newblock \showarticletitle{The complexity of optimal queueing network control}. In \bibinfo{booktitle}{\emph{Proceedings of {IEEE} 9th {Annual} {Conference} on {Structure} in {Complexity} {Theory}}}. \bibinfo{pages}{318--322}.
\newblock
\urldef\tempurl%
\url{https://doi.org/10.1109/SCT.1994.315792}
\showDOI{\tempurl}


\bibitem[Papadimitriou and Tsitsiklis(1999)]%
        {papadimitriou_complexity_1999}
\bibfield{author}{\bibinfo{person}{Christos~H. Papadimitriou} {and} \bibinfo{person}{John~N. Tsitsiklis}.} \bibinfo{year}{1999}\natexlab{}.
\newblock \showarticletitle{The {Complexity} of {Optimal} {Queuing} {Network} {Control}}.
\newblock \bibinfo{journal}{\emph{Mathematics of Operations Research}} \bibinfo{volume}{24}, \bibinfo{number}{2} (\bibinfo{year}{1999}), \bibinfo{pages}{293--305}.
\newblock
\showISSN{0364-765X}
\newblock
\shownote{Publisher: INFORMS}.


\bibitem[Powell(2011)]%
        {cochran_knowledge_2011}
\bibfield{author}{\bibinfo{person}{Warren~B. Powell}.} \bibinfo{year}{2011}\natexlab{}.
\newblock \showarticletitle{The {Knowledge} {Gradient} for {Optimal} {Learning}}.
\newblock In \bibinfo{booktitle}{\emph{Wiley {Encyclopedia} of {Operations} {Research} and {Management} {Science}}}. \bibinfo{publisher}{John Wiley \& Sons, Inc.}, \bibinfo{address}{Hoboken, NJ, USA}.
\newblock
\showISBNx{978-0-470-40053-1}
\urldef\tempurl%
\url{https://doi.org/10.1002/9780470400531.eorms0444}
\showDOI{\tempurl}


\bibitem[Richardson and Gilbert(2021)]%
        {richardson_framework_2021}
\bibfield{author}{\bibinfo{person}{Brianna Richardson} {and} \bibinfo{person}{Juan~E. Gilbert}.} \bibinfo{year}{2021}\natexlab{}.
\newblock \bibinfo{title}{A {Framework} for {Fairness}: {A} {Systematic} {Review} of {Existing} {Fair} {AI} {Solutions}}.
\newblock
\newblock
\urldef\tempurl%
\url{http://arxiv.org/abs/2112.05700}
\showURL{%
\tempurl}
\newblock
\shownote{arXiv:2112.05700 [cs]}.


\bibitem[Shi et~al\mbox{.}(2022)]%
        {shi_sampling_2022}
\bibfield{author}{\bibinfo{person}{Jiaxin Shi}, \bibinfo{person}{Chang Liu}, {and} \bibinfo{person}{Lester Mackey}.} \bibinfo{year}{2022}\natexlab{}.
\newblock \bibinfo{title}{Sampling with {Mirrored} {Stein} {Operators}}.
\newblock
\newblock
\urldef\tempurl%
\url{http://arxiv.org/abs/2106.12506}
\showURL{%
\tempurl}
\newblock
\shownote{arXiv:2106.12506 [cs, stat]}.


\bibitem[Shi and Mackey(2023)]%
        {shi_finite-particle_2023}
\bibfield{author}{\bibinfo{person}{Jiaxin Shi} {and} \bibinfo{person}{Lester Mackey}.} \bibinfo{year}{2023}\natexlab{}.
\newblock \bibinfo{title}{A {Finite}-{Particle} {Convergence} {Rate} for {Stein} {Variational} {Gradient} {Descent}}.
\newblock
\newblock
\urldef\tempurl%
\url{http://arxiv.org/abs/2211.09721}
\showURL{%
\tempurl}
\newblock
\shownote{arXiv:2211.09721 [cs, stat]}.


\bibitem[Shi et~al\mbox{.}(2020)]%
        {shi_artificial_2020}
\bibfield{author}{\bibinfo{person}{Zheyuan~Ryan Shi}, \bibinfo{person}{Claire Wang}, {and} \bibinfo{person}{Fei Fang}.} \bibinfo{year}{2020}\natexlab{}.
\newblock \bibinfo{title}{Artificial {Intelligence} for {Social} {Good}: {A} {Survey}}.
\newblock
\newblock
\urldef\tempurl%
\url{http://arxiv.org/abs/2001.01818}
\showURL{%
\tempurl}
\newblock
\shownote{arXiv:2001.01818 [cs]}.


\bibitem[Silver et~al\mbox{.}(2018)]%
        {silver_general_2018}
\bibfield{author}{\bibinfo{person}{David Silver}, \bibinfo{person}{Thomas Hubert}, \bibinfo{person}{Julian Schrittwieser}, \bibinfo{person}{Ioannis Antonoglou}, \bibinfo{person}{Matthew Lai}, \bibinfo{person}{Arthur Guez}, \bibinfo{person}{Marc Lanctot}, \bibinfo{person}{Laurent Sifre}, \bibinfo{person}{Dharshan Kumaran}, \bibinfo{person}{Thore Graepel}, \bibinfo{person}{Timothy Lillicrap}, \bibinfo{person}{Karen Simonyan}, {and} \bibinfo{person}{Demis Hassabis}.} \bibinfo{year}{2018}\natexlab{}.
\newblock \showarticletitle{A general reinforcement learning algorithm that masters chess, shogi, and {Go} through self-play}.
\newblock \bibinfo{journal}{\emph{Science}} \bibinfo{volume}{362}, \bibinfo{number}{6419} (\bibinfo{date}{Dec.} \bibinfo{year}{2018}), \bibinfo{pages}{1140--1144}.
\newblock
\showISSN{0036-8075, 1095-9203}
\urldef\tempurl%
\url{https://doi.org/10.1126/science.aar6404}
\showDOI{\tempurl}


\bibitem[Sombabu et~al\mbox{.}(2020)]%
        {sombabu_whittle_2020}
\bibfield{author}{\bibinfo{person}{Bejjipuram Sombabu}, \bibinfo{person}{Aditya Mate}, \bibinfo{person}{D. Manjunath}, {and} \bibinfo{person}{Sharayu Moharir}.} \bibinfo{year}{2020}\natexlab{}.
\newblock \showarticletitle{Whittle {Index} for {AoI}-{Aware} {Scheduling}}. In \bibinfo{booktitle}{\emph{2020 {International} {Conference} on {COMmunication} {Systems} {NETworkS} ({COMSNETS})}}. \bibinfo{pages}{630--633}.
\newblock
\urldef\tempurl%
\url{https://doi.org/10.1109/COMSNETS48256.2020.9027444}
\showDOI{\tempurl}
\newblock
\shownote{ISSN: 2155-2509}.


\bibitem[Sutton and Barto(2018)]%
        {sutton_reinforcement_2018}
\bibfield{author}{\bibinfo{person}{Richard~S. Sutton} {and} \bibinfo{person}{Andrew~G. Barto}.} \bibinfo{year}{2018}\natexlab{}.
\newblock \bibinfo{booktitle}{\emph{Reinforcement learning: an introduction} (\bibinfo{edition}{second edition} ed.)}.
\newblock \bibinfo{publisher}{The MIT Press}, \bibinfo{address}{Cambridge, Massachusetts}.
\newblock
\showISBNx{978-0-262-03924-6}


\bibitem[Varshney and Mojsilovic(2019)]%
        {varshney_open_2019}
\bibfield{author}{\bibinfo{person}{Kush~R. Varshney} {and} \bibinfo{person}{Aleksandra Mojsilovic}.} \bibinfo{year}{2019}\natexlab{}.
\newblock \bibinfo{title}{Open {Platforms} for {Artificial} {Intelligence} for {Social} {Good}: {Common} {Patterns} as a {Pathway} to {True} {Impact}}.
\newblock
\newblock
\urldef\tempurl%
\url{http://arxiv.org/abs/1905.11519}
\showURL{%
\tempurl}
\newblock
\shownote{arXiv:1905.11519 [cs]}.


\bibitem[Wang et~al\mbox{.}(2012)]%
        {wang_optimality_2012}
\bibfield{author}{\bibinfo{person}{Kehao Wang}, \bibinfo{person}{Lin Chen}, \bibinfo{person}{Quan Liu}, {and} \bibinfo{person}{Khaldoun~Al Agha}.} \bibinfo{year}{2012}\natexlab{}.
\newblock \showarticletitle{On {Optimality} of {Myopic} {Policy} for {Restless} {Multi}-armed {Bandit} {Problem} with {Non} i.i.d. {Arms} and {Imperfect} {Detection}}.
\newblock \bibinfo{journal}{\emph{IEEE Transactions on Signal Processing}} \bibinfo{volume}{60}, \bibinfo{number}{1} (\bibinfo{date}{Jan.} \bibinfo{year}{2012}), \bibinfo{pages}{300--309}.
\newblock
\showISSN{1053-587X, 1941-0476}
\urldef\tempurl%
\url{https://doi.org/10.1109/TSP.2011.2170684}
\showDOI{\tempurl}
\newblock
\shownote{arXiv:1205.5375 [cs]}.


\bibitem[Wang et~al\mbox{.}(2022a)]%
        {wang_optimistic_2022}
\bibfield{author}{\bibinfo{person}{Kai Wang}, \bibinfo{person}{Lily Xu}, \bibinfo{person}{Aparna Taneja}, {and} \bibinfo{person}{Milind Tambe}.} \bibinfo{year}{2022}\natexlab{a}.
\newblock \bibinfo{title}{Optimistic {Whittle} {Index} {Policy}: {Online} {Learning} for {Restless} {Bandits}}.
\newblock
\newblock
\urldef\tempurl%
\url{http://arxiv.org/abs/2205.15372}
\showURL{%
\tempurl}
\newblock
\shownote{arXiv:2205.15372 [cs]}.


\bibitem[Wang et~al\mbox{.}(2022b)]%
        {wang_optimal_2022}
\bibfield{author}{\bibinfo{person}{Kehao Wang}, \bibinfo{person}{Jihong Yu}, \bibinfo{person}{Lin Chen}, \bibinfo{person}{Pan Zhou}, {and} \bibinfo{person}{Moe~Z. Win}.} \bibinfo{year}{2022}\natexlab{b}.
\newblock \showarticletitle{Optimal {Myopic} {Policy} for {Restless} {Bandit}: {A} {Perspective} of {Eigendecomposition}}.
\newblock \bibinfo{journal}{\emph{IEEE Journal of Selected Topics in Signal Processing}} \bibinfo{volume}{16}, \bibinfo{number}{3} (\bibinfo{date}{April} \bibinfo{year}{2022}), \bibinfo{pages}{420--433}.
\newblock
\showISSN{1941-0484}
\urldef\tempurl%
\url{https://doi.org/10.1109/JSTSP.2022.3142502}
\showDOI{\tempurl}
\newblock
\shownote{Conference Name: IEEE Journal of Selected Topics in Signal Processing}.


\bibitem[Weber and Weiss(1990)]%
        {weber_index_1990}
\bibfield{author}{\bibinfo{person}{Richard~R. Weber} {and} \bibinfo{person}{Gideon Weiss}.} \bibinfo{year}{1990}\natexlab{}.
\newblock \showarticletitle{{ON} {AN} {INDEX} {POLICY} {FOR} {RESTLESS} {BANDITS}}.
\newblock \bibinfo{journal}{\emph{Journal of Applied Probability}} \bibinfo{volume}{27}, \bibinfo{number}{3} (\bibinfo{year}{1990}), \bibinfo{pages}{637--648}.
\newblock


\bibitem[Whittle(1988)]%
        {whittle_restless_1988}
\bibfield{author}{\bibinfo{person}{P. Whittle}.} \bibinfo{year}{1988}\natexlab{}.
\newblock \showarticletitle{Restless {Bandits}: {Activity} {Allocation} in a {Changing} {World}}.
\newblock \bibinfo{journal}{\emph{Journal of Applied Probability}}  \bibinfo{volume}{25} (\bibinfo{year}{1988}), \bibinfo{pages}{287--298}.
\newblock
\showISSN{0021-9002}
\urldef\tempurl%
\url{https://doi.org/10.2307/3214163}
\showDOI{\tempurl}
\newblock
\shownote{Publisher: Applied Probability Trust}.


\bibitem[Xiong et~al\mbox{.}(2022a)]%
        {xiong_index-aware_2022}
\bibfield{author}{\bibinfo{person}{Guojun Xiong}, \bibinfo{person}{Xudong Qin}, \bibinfo{person}{Bin Li}, \bibinfo{person}{Rahul Singh}, {and} \bibinfo{person}{Jian Li}.} \bibinfo{year}{2022}\natexlab{a}.
\newblock \showarticletitle{Index-aware reinforcement learning for adaptive video streaming at the wireless edge}. In \bibinfo{booktitle}{\emph{Proceedings of the {Twenty}-{Third} {International} {Symposium} on {Theory}, {Algorithmic} {Foundations}, and {Protocol} {Design} for {Mobile} {Networks} and {Mobile} {Computing}}} \emph{(\bibinfo{series}{{MobiHoc} '22})}. \bibinfo{publisher}{Association for Computing Machinery}, \bibinfo{address}{New York, NY, USA}, \bibinfo{pages}{81--90}.
\newblock
\showISBNx{978-1-4503-9165-8}
\urldef\tempurl%
\url{https://doi.org/10.1145/3492866.3549726}
\showDOI{\tempurl}


\bibitem[Xiong et~al\mbox{.}(2022b)]%
        {xiong_model-free_2022}
\bibfield{author}{\bibinfo{person}{Guojun Xiong}, \bibinfo{person}{Shufan Wang}, \bibinfo{person}{Jian Li}, {and} \bibinfo{person}{Rahul Singh}.} \bibinfo{year}{2022}\natexlab{b}.
\newblock \bibinfo{title}{Model-free {Reinforcement} {Learning} for {Content} {Caching} at the {Wireless} {Edge} via {Restless} {Bandits}}.
\newblock
\newblock
\urldef\tempurl%
\url{http://arxiv.org/abs/2202.13187}
\showURL{%
\tempurl}
\newblock
\shownote{arXiv:2202.13187 [cs]}.


\bibitem[Zhang(2018)]%
        {zhang_improved_2018}
\bibfield{author}{\bibinfo{person}{Zijun Zhang}.} \bibinfo{year}{2018}\natexlab{}.
\newblock \showarticletitle{Improved {Adam} {Optimizer} for {Deep} {Neural} {Networks}}. In \bibinfo{booktitle}{\emph{2018 {IEEE}/{ACM} 26th {International} {Symposium} on {Quality} of {Service} ({IWQoS})}}. \bibinfo{pages}{1--2}.
\newblock
\urldef\tempurl%
\url{https://doi.org/10.1109/IWQoS.2018.8624183}
\showDOI{\tempurl}
\newblock
\shownote{ISSN: 1548-615X}.


\end{thebibliography}

\appendix

\section{Related Technical Work and Discussion}\label{sec:rel_work}

We review RMAB algorithms when transition probabilities are completely unknown. We distinguish between approaches intended for fully observed and partially observed state-to-state transitions. Finally, we identify the gap in the literature that our algorithm fills: partially monitored, episodic, online RMABs with unobserved state transition labels for finitely many states and any deterministic policy. 

While the RMAB framework conceptually fits well in stochastic scheduling applications, the original RMAB formulation's planning problem is impossible to solve directly with normal reinforcement learning (RL) methods, e.g., \citet{papadimitriou_complexity_1994, papadimitriou_complexity_1999}, and therefore admits two challenging subproblems for its application. The first is how to choose the highest-reward arms at each timestep over a combinatorial state and action space, and is referred to as the ``planning problem.''  Although the planning problem is PSPACE-hard in general, i.e. intractable to compute, \citet{whittle_restless_1988} elegantly works around this issue by replacing the RMAB's strict budget constraint on $M$ at \textit{every} time step with a constraint that holds across time \textit{on average}. Conceptually, Whittle accomplishes this by recasting the constrained RMAB problem as a Lagrangian optimization problem among decoupled arms \citep{whittle_restless_1988}. \citet{whittle_restless_1988} shows that if the new budget constraint holds on average rather than for all time steps, one can solve an auxiliary planning problem, and the cumulative reward from solving it upper-bounds the original RMAB objective. (See Appendix~\ref{wrelax} for a precise description of Whittle's relaxation.)  \citet{whittle_restless_1988}'s popular heuristic method for solving the auxiliary planning problem is called the Whittle Index policy (WIP). A WIP selects the top $M$ best arms each time step based on an ordinal ranking of arms according to a state-dependent value called the Whittle index (WI). And, if there exists an ordinal ranking of the arms via Whittle indices, we say that the RMAB is \textit{indexable}. Crucially, the Whittle policy is \textit{only} optimal when an RMAB is indexable, but verifying indexability is nontrivial and computing Whittle indices is computationally hard \citep{whittle_restless_1988, papadimitriou_complexity_1994}. Nonetheless, Whittle policies stand in as a gold standard because they are heuristically intuitive, perform well in practice, and have optimality results for some cases, e.g., \citep{weber_index_1990, hodge_asymptotic_nodate}. 

Due to the difficulty of the RMAB problem even under \citet{whittle_restless_1988}'s relaxation, a significant amount of work explores ways to bypass estimating transition matrices altogether. The approach is the following. Even when transition dynamics are known, one must devise a way to estimate Whittle indices. However, this is nontrivial, as \citet{whittle_restless_1988} did not offer a generalizable definition of Whittle indices. Even if an estimation procedure is determined, often the computational problem presents significant issues in its own right \citep{mate_collapsing_nodate}. Further, these concerns exist for small-to-moderate state spaces for this literature, e.g., $2 \leq|\State| \leq 4$, and RMAB models for practical applications may face much larger state spaces than $|\State| = 4$ . Therefore, special structure can be essential for emulating and extending \citet{whittle_restless_1988}'s approach, e.g., the existence of optimal threshold policies for decoupled DTMCs or an explicit Whittle index formula \citep{borkar_reinforcement_2018, avrachenkov_learning_2019}. Unfortunately, relying on the existence of special structure limits the generality of such results and works against our larger objective of finding non-bespoke solutions across social good organizations. Consequently, to achieve our goals, we can divide the literature into two categories: \textit{offline planning} for RMABs versus \textit{online learning} for RMABs. As the current work falls under the latter domain, we briefly touch on the former before focusing on the latter. 

In offline planning for RMABs, algorithms avoid estimating transition dynamics in real-time.\footnote{While there are other offline planning algorithms that start by assuming transition dynamics are known, these are not relevant for our work, so we focus on cases where transition dynamics are unknown and estimation is bypassed.} This reduces the RMAB to a sequential decision problem where the goal is simply to produce the best plan (or policy) for sequentially maximizing a cumulative reward objective or minimizing a cumulative regret objective. In this area of work, the literature turned to exploring $Q$-learning methods for RMABs due to its history, rich literature, and recent major successes in the RL community \citep{mnih_human-level_2015, silver_general_2018}.

In contrast to offline planning, the online learning setting in RMABs (and in RL, generally) refers to when the transition dynamics are unknown and must be learned (or estimated) in real-time as the algorithm interacts with the environment. In this case, the RMAB must simultaneously estimate transition probabilities while gathering data about the state space, often referred to as ``exploration'' in RL, while balancing the problem of choosing actions that maximize an objective function (e.g., minimize regret for maximize reward) with respect to some benchmark, often referred to as ``exploitation'' in RL. However, in addition to estimating transition matrices well, researchers in the online learning branch of the RMAB literature need to handle two other issues. First, if the RMAB is not indexable, then they need to find a more general solution without this extra problem structure. Second, they need this solution to be computationally tractable for moderate-to-large state spaces in RMABs, e.g., $|\State| > 4$. 

One approach extends upper confidence bound (UCB) methods from the classic bandit literature to RMABs. Broadly speaking, UCB methods refer to estimating confidence sets over the reward distribution for each arm and then selecting arm(s) by taking an argmax over all of the arms' upper confidence estimates \citep{lai_asymptotically_1985, auer_using_2003, sutton_reinforcement_2018, lattimore_bandit_2020}. \citet{wang_optimistic_2022}, for example, restrict its benchmark policy to the class of Whittle index policies and leverage the weak-decomposability of the WIP. \citet{wang_optimistic_2022}'s method, however, is likely to be too computationally costly to validate in practice, as their method would require estimating, maintaining, and validating a confidence set per arm over transition parameters. 

Further, costly evaluations for confidence sets may also lead to statistical inefficiencies when approximation is needed. \textit{This is always the case for feasible UCB solutions in RMABs} because, relative to Thompson sampling (TS) approaches, UCB approaches need costly evaluations for confidence sets that may also suffer from statistical inefficiencies when approximation is needed \citep{osband_why_2017}. At a high level, this is because UCB approaches follow the \textit{optimism in the face of uncertainty} (OFU) principle, which ultimately produces \textit{rectangular} rather than \textit{elliptical} confidence sets over MDP parameters \citep{osband_why_2017}. \citet{osband_why_2017} show that the corners of the confidence set constructed by UCB-style algorithms will be $\sqrt{\vert S \vert}$ misspecified to the underlying true (elliptical) distribution when combined across $\vert S \vert$ independent estimates. While there are algorithms that address this discrepancy, their worst case performance ultimately switches from growing as a function of the state space's size to growing as a function of the time horizon's length, which is equally problematic \citep{osband_why_2017}. Returning to RMABs, the loose confidence set issue is likely to be exacerbated by the number of arms and combinatorial state space, depending on the method, which may cause further loss of statistical efficiency, yet \citet{wang_optimistic_2022} offer no formal guarantees or empirical examinations into the statistical efficiency of their method.

Finally, \citet{wang_optimistic_2022} is only designed for fully observed RMABs, i.e, each arm's state is observed each time step, and unknown transition probabilities. Nonetheless, while \citet{wang_optimistic_2022}'s method works for the online learning problem in RMABs, it cannot accommodate partial observability \textit{and} $k$-step transitions induced by partial monitoring of the arms (see Section \ref{subsec:learning}). 

Outside of UCB methods, TS takes a posterior sampling approach to estimate transition dynamics. Specifically, each arm is assigned a prior distribution over transition parameters and the RMAB plays a best-response action with respect to a sampled set of transition paramters. Conditional on the observed state, action, and reward, the selected arms receive a posterior update \citep{lattimore_bandit_2020}. Among TS-based methods for RMABs, \citet{jung_regret_2019} design an algorithm for the episodic, finite-horizon case when an RMAB has only two states and actions. \citet{jung_thompson_2019} extend their work to the non-episodic, infinite horizon setting by adapting the Thompson sampling for Dynamic Episodes (TSDE) algorithm from the RL literature to the RMAB setting \citep{ouyang_learning_2017}. \citet{akbarzadeh_learning_2022} also adapt TSDE to the infinite horizon setting, but \citet{jung_thompson_2019} work in the partially observed case whereas \citet{akbarzadeh_learning_2022} works in either the fully or partially observed case. \citet{jung_regret_2019} and \citet{jung_thompson_2019}, however, work in a more general policy setting that flexibly and smoothly allows the user to apply any deterministic policy used in an RMAB. In contrast, \citet{akbarzadeh_learning_2022} specializes their analysis to directly rely on structure from the WI and WIP. Importantly, all prior work assumes convenient posterior updates, e.g., conjugate sampling, while we explicitly allow for approximate Bayesian inference in our TS algorithm.

\section{Supplementary Results}

\subsection{ Quality of $k$-Step Estimates}\label{supp:trans_sens_kStep}

This section demonstrates how our estimation procedure performs under various levels of $k$ for unlabeled transitions. We study the case of single-particle MSVGD. We start our simulation procedure by sampling a true underlying set of Dirichlet parameters for the $\vert \State \vert \times \vert \State \vert$ transition parameters. From there, at the start of each simulation run we draw a random sample of Dirichlet parameters from a $\text{Dir}( \textbf{5}^{\vert \State \vert})$ distribution. Let $n_s$ denote the number of SVGD iterations. For each combination of $T$, $k$, $n_{s}$, and $\vert \State \vert$, we run $30$ replications. For each replicate, we measure the quality of the final estimates, by the KL divergence (KLD) between $\hat{\boldsymbol{\alpha}}$ and $\boldsymbol{\alpha}^{\star}$. 

\paragraph{KL Divergence for $1$-particle MSVGD by $T$ and $k$ for $\vert \State \vert = 10$.}

In this study we consider the following levels for each factor: $T \in [50, 100, 200]$, $\vert \State \vert = 10$, $k \in [1,5,10]1$, $n_{s} = 40$. Across all analyses, our extension of \citet{kohjima_learning_2020}'s MC recovery algorithm combined with finite-particle SVGD reports KLD estimates close to $0$. This suggests that our estimated parameters $\hat{\boldsymbol{\alpha}}$ closely approximate the true underlying Dirichlet parameters $\boldsymbol{\alpha}^{\star}$ across reported levels of $T$ and $k$.

\begin{figure}[!ht]
     \centering
         \centering
         \includegraphics[width=.5\linewidth, height =.4\linewidth]{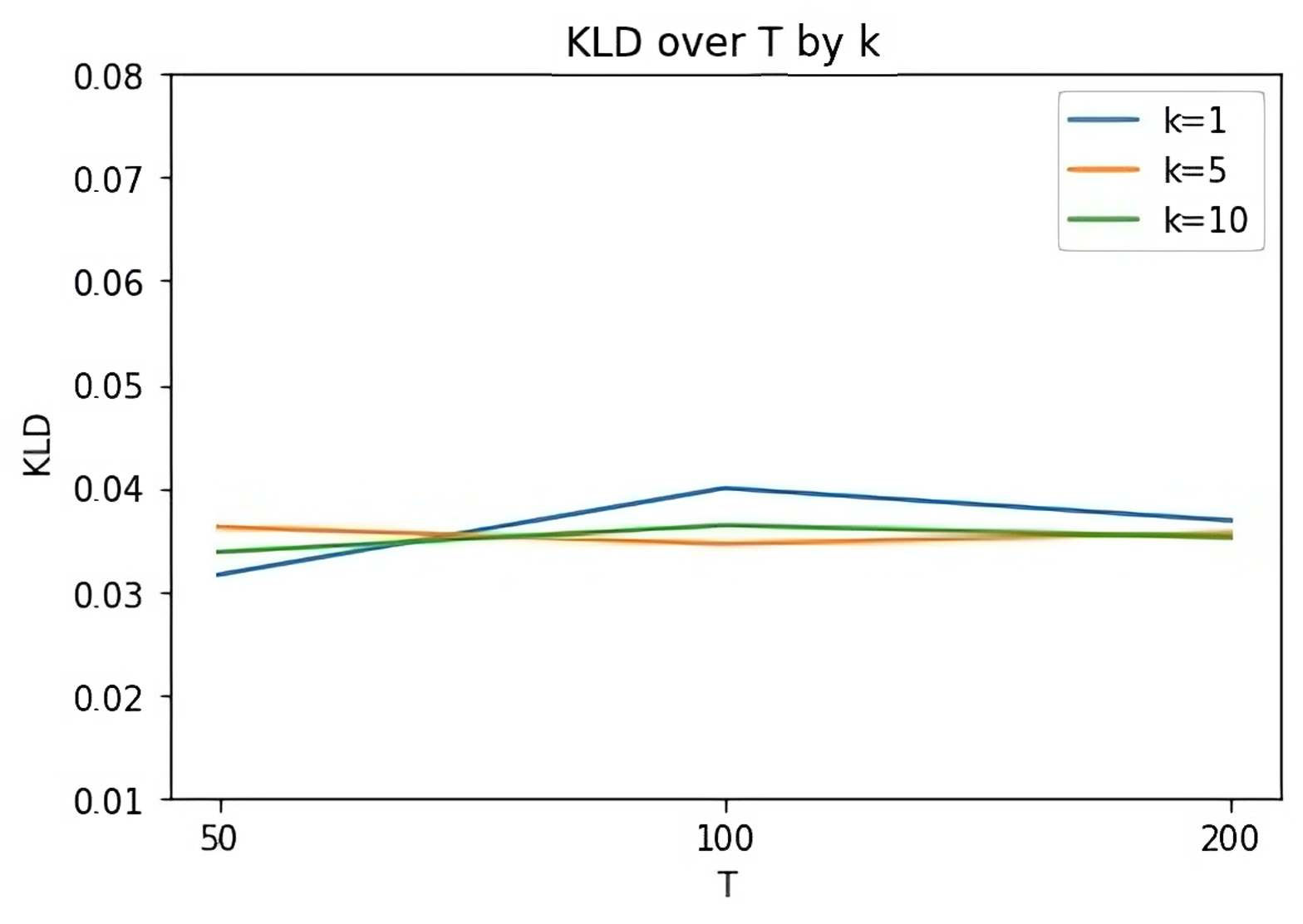}
         
    \caption{Average $KLD(\hat{\boldsymbol{\alpha}} \vert \boldsymbol{\alpha}^{\star})$ using MSVGD using MSVGD with $k$-step transition estimation. $T \in [50, 100, 200]$, $\vert \State \vert = 10$, $k \in [1,5,10]1$, $n_{s} = 40$.}
    \label{fig:kld_over_k}
\end{figure}

\subsection{Quality of $\eta$ Estimations}\label{supp:trans_sens_eta}

This section shows how the $k$-step estimation component of our estimation procedure, $f(\cdot \vert \eta)$, performs under various hyperparameter settings. While one can work with any exponential family model $f$, we study the Poisson model, as this is what we use in Section \ref{single_particle:msvgd}.

We start our simulation procedure by sampling a true underlying set of Dirichlet parameters for the $\vert \State \vert \times \vert \State \vert$ transition parameters. From there, at the start of each simulation run we draw a random sample of Dirichlet parameters from a $\text{Dir}( \textbf{5}^{\vert \State \vert})$ distribution. Let $n_s$ denote the number of SVGD iterations. For each combination of $T$, $k$, $n_{s}$, and $\vert \State \vert$, we run $30$ replications. For each replicate, we measure the quality of the final estimates by the MAE between $\hat{\boldsymbol{\eta}}$ and $\boldsymbol{\eta}^{\star}$. 

\begin{figure}[!ht]
     \centering
         \includegraphics[width=.5\linewidth, height =.4\linewidth]{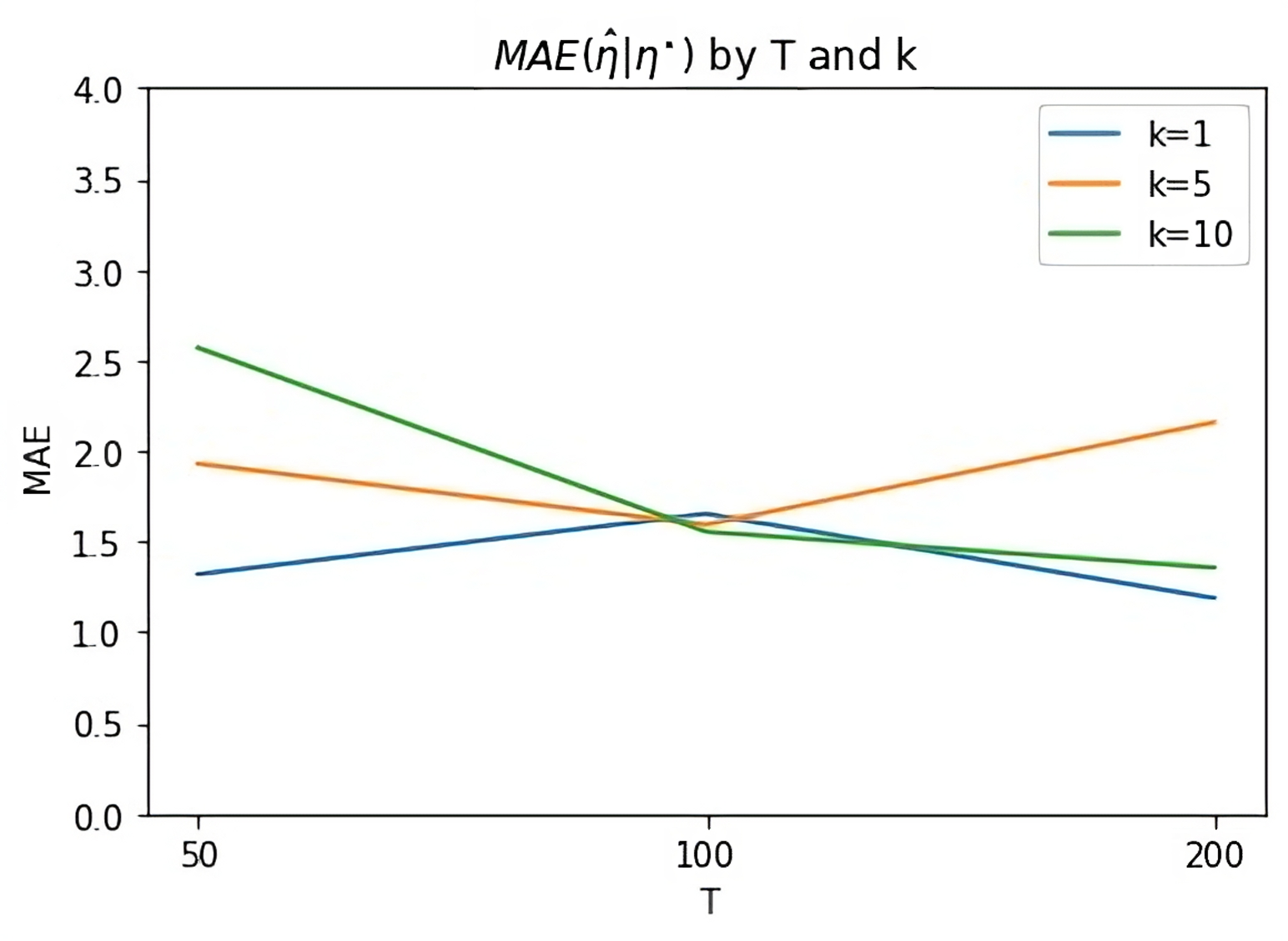}
    \caption{Average $MAE(\hat{\boldsymbol{\eta}} \vert \boldsymbol{\eta}^{\star})$ using MSVGD with $k$-step transition estimation. $T \in [50, 100, 200]$, $\vert \State \vert = 10$, $k \in [1,5,10]1$, $n_{s} = 40$.}
    \label{fig:mae_over_k}
\end{figure}

\paragraph{MAE for $1$-particle MSVGD by $T$ and $k$ for $\vert \State \vert = 12$.} We consider the following levels for each factor: $T \in [50, 100, 200]$, $\vert \State \vert = 12$, $k \in [1,5,10]$, $n_{s} = 40$. Across all analyses in Figure \ref{fig:mae_over_k}, our extension of \citet{kohjima_learning_2020}'s MC recovery algorithm combined with finite-particle MSVGD reports small MAE estimates, indicating that our final estimates approach the true underlying parameters across levels of $T$ and $k$.

\section{Background}

\subsection{Stein Variational Gradient Descent}\label{app:svgd}

Let $k_x(\cdot) = k(x,\cdot)$ be a positive definite (p.d.) kernel in a reproducing kernel Hilbert space (RKHS) \footnote{SVGD depends on the theory of reproducing kernel Hilbert spaces (RKHS), so we offer a brief review in Section \ref{supp:rkhs}.} and denote the score function of the target distribution as $\nabla_{\theta}\log p(\theta) := s(\theta)$. Classic SVGD approximates the target density $p$ by a sequence of empirical densities $q^{\ell}$ as follows. Given an initial sample of parameters drawn from some prior $q_0$, denoted $(x_{i}^{0})_{i \in [n]} := \sum_{i \in [n]} \delta_{x_i}$,

\begin{displaymath}
    x_{i}^{\ell + 1} \xleftarrow[]{} x_{i}^{\ell} + \epsilon^{\ell} \widehat{\phi(x^{\ell}_{i})}, 
\end{displaymath}
\noindent where 
\begin{displaymath}
    \widehat{\phi(x^{\ell}_{i})} \xleftarrow[]{} \frac{1}{n}\sum_{j=1}{k(x^{\ell}_{j}, x^{\ell}_{i})s(x_{j}^{\ell}) + \nabla_{x_{j}^{\ell}}k(x^{\ell}_{j}, x^{\ell}_{i})}.
\end{displaymath}

\noindent By conditioning on the available data and transporting an initial set of parameters, or \textit{particles} $(x_{i}^{\ell +1})_{i \in [n]}$, the resulting $(x_{i}^{\ell +1})_{i \in [n]}$ act as the $\ell + 1\text{st}$ approximation, $q^{\ell+1}$, to the target density, $p$.

\subsection{Reproducing Kernel Hilbert Spaces}\label{supp:rkhs}

Let $\Hist$ be a Hilbert space of functions defined on $\mathbb{X}$ (usually a metric space) and taking values in $\mathbb{R}$. $k$ is referred to as a reproducing kernel (or Mercer kernel) of $\Hist$ if for every $x \in \mathbb{X}$, $k(x, \cdot) \in \Hist$ and for every $f \in \Hist$, $\langle f, k(x, \cdot) \rangle_{\Hist} = f(x)$. We call $\Hist$ a reproducing kernel Hilbert space (RKHS) if it has such a kernel. Kernels must be positive definite (p.d.) functions, i.e., any matrix of the form $(k(x_i, x_j))_{ij}$ is positive semi-definite. 

For any p.d. function $k$, there exists an unique RKHS with that $k$ as its reproducing kernel. In particular, this can be constructed via the completion of $\{\sum_{i=1}^{n} a_{i}k(x_i, \cdot)): x_i \in \mathbb{X}, a_i \in \mathbb{R}, i \in \mathbb{N} \}$ \citep{shi_sampling_2022}. For our setting, we assume $\mathbb{X}$ is a metric space, $k$ is a continuous and bounded p.d. kernel associated with RKHS $\Hist$. 

\subsection{Whittle's Relaxation}\label{wrelax}

We formally state \citet{whittle_restless_1988}'s auxiliary problem, called Whittle's relaxation, and formalize the concept of indexability.

One reason the RMAB problem is hard is because the state space increases exponentially in the number of arms, $N$. In fact, the RMAB optimization problem has been shown to be PSPACE-Hard by \citet{papadimitriou_complexity_1994}, i.e., this problem is provably intractable. \citet{whittle_restless_1988}'s derives a solution to this bottleneck in three main steps. First, he relaxes the budget constraint so that it holds on average. Second, under the relaxed budget constraint he reformulates the original RMAB problem as a linear program (LP). Finally, \citet{whittle_restless_1988} rewrites the LP as a Lagrangian optimization problem which can be solved as $N$ independent Lagrange optimization problems.

In \citet{whittle_restless_1988}'s first step, he proposes to change the budget constraint from $\sum_{i \in [N]} A^{i}_{t} = M \hspace{1mm} (\forall t \in \Nz)$ to $\frac{1}{T} \underset{T \xrightarrow{} \infty}{\liminf}\hspace{1mm}E(\sum_{t=0}^{T-1}\sum_{i \in [N]} A^{i}_{t}) = M$, relaxing the strict constraint so that only $M$ projects need to be active on average. As a result, the optimal solution to the relaxed RMAB problem forms an upper bound on any solution to the original RMAB problem. Denote the average reward for the original problem by $\Bar{R}$ and the average reward for the auxiliary problem associated with the relaxed constraint by $\Bar{R}_{\text{relax}}$. For any solution to the relaxed RMAB problem, we have that $\Bar{R} \leq \Bar{R}_{\text{relax}}$. To see this, observe that any optimal policy activates as many arms as possible per time-step, due to the monotonicity assumption. If the only policies being considered are Whittle policies, then any optimal Whittle policy's performance is monotonically increasing as a function of the budget size $M$. Therefore the average reward from the relaxed RMAB problem is always at least the average reward of the original problem, because the relaxed budget constraint is at least as large as the strict budget constraint $\forall t \in \Nz$.

Next, \cite{whittle_restless_1988} exploits the relaxed budget constraint to write the auxiliary RMAB problem as a tractable LP, based on a standard LP formulation of time-average MDPs (\cite{manne_linear_1960}). Denote the state-action visitation frequency under a policy $\pi$ by $g(x,a) = \lim_{T\xrightarrow[]{}\infty} \frac{1}{T}\mathbb{E}_{\pi}(\sum_{t=0}^{T-1} \ind(A_t=a, X_t = x))$. This is a standard state-action frequency measure representing the expected long-run fraction of time that action $a$ is taken under policy $\pi$ in state $s$. Denote $\mathcal{U}_i$ as the feasible set for $g$, i.e., the bounded polytope imposed by LP constraints for arm $i$'s MDP (\cite{manne_linear_1960}). Then the auxiliary problem can be written as the following LP (see \cite{nino-mora_restless_2001} for details).

\[
\max \sum_{x \in \State} r(x, 0) g(x,0) + \sum_{x \in \State} r(x, 1) g(x,1)
\]
\[
\text{s.t.}\hspace{2mm} \sum_{x \in \State} g(x,1) = M, \hspace{3mm}  \sum_{x \in \State}g(x,0) + g(x,1) = N, \hspace{1mm} \text{and} 
\]
\[
\Bigg(g(y,0), g(y,1)\Bigg)_{y \in \State_i} \in \mathcal{U}_i, \hspace{2mm} (i \in [N])
\]
\noindent
where the second constraint is implicit. As an LP, the auxiliary RMAB problem is now solvable in polynomial time by LP interior point algorithms (\cite{nino-mora_restless_2001}). At this point \citet{whittle_restless_1988} dualizes the constraints and introduces the Lagrange multiplier $\lambda$ to obtain the Lagrangian relaxation,

\[
\Ell(\lambda) = \sum_{x \in \State} (r(x, 0)+\lambda) g(x,0) + \sum_{x \in \State} r(x, 1) g(x,1) - \lambda(N-M).
\]
\[
s.t. \hspace{3mm} \Bigg(g(y,0), g(y,1)\Bigg)_{y \in \State_i} \in \mathcal{U}_i, \hspace{2mm} (i \in [N])
\]

\noindent
Note that from the first summand in this equation, we can intuitively view $\lambda$ as a subsidy for the passive action. Recognizing that the cost and constraint are separable, the Lagrangian relaxation allows for optimization of $N$ individual control problems. For a single arm $i \in [N]$, the problem is now to maximize

\[
\Ell_{i}(\lambda) = \sum_{x \in \State_i} (r(x, 0)+\lambda) g(x,0) + \sum_{x \in \State_i} r(x, 1) g(x,1),
\]
\[
s.t. \hspace{3mm} \Bigg(g(y,0), g(y,1)\Bigg)_{y \in \State_i} \in \mathcal{U}_i, \hspace{2mm} (i \in [N])
\]
\noindent
Therefore the fully decoupled Lagrange optimization problem for the auxiliary RMAB problem is to maximize

\[
\Ell(\lambda) = \sum_{i \in [N]} \Ell_{i}(\lambda) - \lambda(N-M)
\]
\noindent
Finally, we mention another way that some authors represent $\Ell$. By the independence of arms ( \ref{indptassn}), we can optimize $\Ell(\theta)$ by optimizing each $\Ell_{i}(\lambda) - \lambda(1-\ind(a=1))$ term arm-by-arm:

\[
\Ell(\lambda) = \sum_{i \in [N]} \Big(\Ell_{i}(\lambda) - \lambda(1-\ind(a=1)) \Big) = \sum_{i \in [N]} \Big(\Ell_{i}(\lambda) - \lambda(1-a) \Big)
\]
\noindent

 In summary, this relaxation allows for each arm's MDP to be solved independently, which effectively forces the state space to grow linearly in $N$ rather than exponentially. The MDPs of the arms are coupled only through the control policy $\pi$ now, based on an ordinal rank-assigning functional of the states known as the Whittle index. 

As we observed, the Whittle index is the Lagrange multiplier, $\lambda$, associated with each arm $i \in [N]$ and can be interpreted as a subsidy for choosing passive arms. In economic terms, the Whittle index for an arm is the subsidy one must pay to make the marginal cost of activating an arm equal to the marginal benefit of activating it. In this way, the Whittle index captures an indirect notion of priority: as the Whittle index increases for an arm, it is more costly to ignore in terms of the opportunity cost of rewards. Finally, the Whittle index is then used to dictate the behavior of the Whittle policy, which selects the top $M$ arms according to their Whittle index ranking. 

As a final remark, we can restate \ref{indassn} more precisely: we now say an arm is \textit{indexable} when $\State_0$  increases monotonically from $\emptyset$ to the $\State$ as $\lambda$ monotonically increases from $-\infty$ to $\infty$. If every arm $i \in [N]$ is indexable, the RMAB is said to be indexable.

\end{document}